\documentclass{article}

\usepackage{microtype}
\usepackage{graphicx}
\usepackage{subcaption}
\usepackage{booktabs} 

\usepackage{hyperref}

\usepackage[T1]{fontenc} 
\usepackage{multirow}
\usepackage{listings}
\usepackage{pifont}
\usepackage{nicematrix,enumitem}

\newcommand{\ie}{\textit{i}.\textit{e}.}
\newcommand{\eg}{\textit{e}.\textit{g}.}
\newcommand{\bftab}{\fontseries{b}\selectfont}
\newcommand{\tsup}[1]{$^\text{#1}$}

\usepackage[accepted]{icml2025}

\usepackage{amsmath}
\usepackage{amssymb}
\usepackage{mathtools}
\usepackage{amsthm}

\usepackage[capitalize,noabbrev]{cleveref}

\theoremstyle{plain}

\theoremstyle{definition}

\theoremstyle{remark}

\usepackage[textsize=tiny]{todonotes}

\icmltitlerunning{GraphGPT: Generative Pre-trained Graph Eulerian Transformer}

\begin{document}

\twocolumn[
\icmltitle{GraphGPT: Generative Pre-trained Graph Eulerian Transformer}



\icmlsetsymbol{equal}{*}

\begin{icmlauthorlist}
\icmlauthor{Qifang Zhao}{comp}
\icmlauthor{Weidong Ren}{comp}
\icmlauthor{Tianyu Li}{comp}
\icmlauthor{Hong Liu}{comp}
\icmlauthor{Xingsheng He}{comp}
\icmlauthor{Xiaoxiao Xu}{comp}
\end{icmlauthorlist}

\icmlaffiliation{comp}{Alibaba Inc., Hangzhou, China}

\icmlcorrespondingauthor{Qifang Zhao}{james.zqf@alibaba-inc.com}
\icmlcorrespondingauthor{Xiaoxiao Xu}{xiaoxiao.xuxx@alibaba-inc.com}

\icmlkeywords{Machine Learning, ICML, GraphGPT, Graph, GPT, BERT, Molecule}

\vskip 0.3in
]



\printAffiliationsAndNotice{}  

\begin{abstract}
We introduce \textit{GraphGPT}, a novel self-supervised \textit{generative pre-trained} model for graph learning based on the \textit{Graph Eulerian Transformer} (\textbf{GET}).
First, we propose GET, which combines a standard transformer encoder or decoder architecture with an innovative graph-to-sequence transformation method.
This method converts graphs or sampled subgraphs into sequences of tokens representing nodes, edges, and attributes in a reversible manner using Eulerian paths.
We pre-train GET using either of the two self-supervised tasks: next-token prediction (NTP) and scheduled masked-token prediction (SMTP). The pre-trained model is then fine-tuned for downstream tasks such as graph-, edge-, and node-level prediction.
Despite its simplicity, GraphGPT achieves performance comparable to or surpassing state-of-the-art methods on multiple large-scale Open Graph Benchmark (OGB) datasets. It demonstrates exceptional results on the molecular property prediction dataset PCQM4Mv2 and the protein-protein interaction dataset ogbl-ppa. Notably, generative pre-training enables scaling GraphGPT to 2 billion parameters while maintaining performance gains — a breakthrough that overcomes the scalability limitations of traditional Graph Neural Networks (GNNs) and prior graph transformers (GTs).
To advance research in graph foundation models and facilitate scientific discovery in chemistry, materials science, and related fields, we have released the source code\footnote{https://github.com/alibaba/graph-gpt} and model checkpoints\footnote{https://www.modelscope.cn/organization/Alibaba-DT}.



\end{abstract}

\section{Introduction}
\label{sec:intro}
The deep learning revolution sparked by AlexNet \citep{alexnet2012} has driven remarkable progress in computer vision (CV) and natural language processing (NLP). The graph learning community similarly shifted from traditional machine learning to deep learning with the rise of graph neural networks (GNNs) \citep{kipf2017semi,hamilton2017inductive,zhang2018link,wu2020comprehensive}.


Today, transformers dominate CV \citep{dosovitskiy2021image,liu2021swin} and NLP \citep{devlin2018bert,radford2018improving}, scaling to billions of parameters \citep{liu2021swin,brown2020language} and achieving superhuman performance on benchmarks like ImageNet \citep{deng2009imagenet} and GLUE \citep{wang2019glue}. These advances underpin transformative applications such as ChatGPT \citep{openai2022chatgpt} and Midjourney \citep{midjourney2023}.

Despite progress, GNNs remain constrained by over-smoothing \citep{rusch2023survey} and over-squashing \citep{alon2021on}, limiting their scalability and capacity to leverage large-scale graph data. Recent efforts to adapt transformers to graphs \citep{ying2021transformers,kim2022pure,luo2022one,muller2023attending} show promise but face critical challenges: 1). \textit{Structural Bias}: Most graph transformers (GTs) rely on handcrafted features or GNN modules to encode graph topology, compromising generalization. 2). \textit{Task Limitations}: GTs excel at graph-level tasks but struggle with edge- and node-level objectives \citep{muller2023attending}. 3). \textit{Pre-Training Gap}: Unlike NLP’s success with self-supervised pre-training \citep{radford2018improving,devlin2018bert}, GTs lack effective frameworks for generative pre-training \citep{min2022transformer,muller2023attending}.




In this work, we propose \textit{GraphGPT}, a novel model for graph learning comprising three key innovations. 1). \textit{GET Backbone}: A transformer-based architecture that operates on graph-equivalent token sequences via Eulerian paths, 2). \textit{Self-Supervised Pre-Training}: Utilizing NTP and SMTP tasks \citep{radford2018improving,chang2022maskedgit}, and 3). \textit{Task-Agnostic Fine-Tuning}: Adapting the pre-trained model to supervised graph-, edge-, and node-level tasks.

Our contributions are summarized as follows:

\begin{itemize}

\item \textbf{Graph Eulerian Transformer (GET)}: We introduce GET, a novel architecture that leverages Eulerian or semi-Eulerian paths\footnote{For a quick recap, a connected graph with every node of even degree is Eulerian, and with exactly two nodes of odd degree is semi-Eulerian. In this paper, `(semi-)Eulerian' refers to both Eulerian and semi-Eulerian unless specified otherwise.} to losslessly and reversibly convert graphs into token sequences. By integrating subgraph sampling and node identity encoding, GET efficiently processes graphs of arbitrary sizes. A standard transformer encoder or decoder is then applied to these sequences, eliminating the need for specialized architectural modifications.

\item \textbf{Generative Pre-Training Framework}: GraphGPT is pre-trained using NTP or SMTP tasks, offering three advantages: $a$) Captures structural and semantic graph patterns without handcrafted features or domain-specific architectures, $b$) Scales to over 2 billion parameters with sustained performance gains, and $c$) Enables effective graph generation through its sequence-based formulation.

\item \textbf{Unified Task Formatting}: We design a novel method to reformat graph-, edge-, and node-level tasks into sequences compatible with transformers. This approach allows downstream tasks to fully exploit pre-trained representations while unifying pretext and target task frameworks.


\item \textbf{State-of-the-Art (SOTA) Performance}: Extensive experiments on OGB datasets demonstrate GraphGPT’s superiority: it achieves SOTA results in graph- and edge-level tasks (e.g., molecular property prediction on PCQM4Mv2 and protein-protein interaction on ogbl-ppa), while delivering competitive performance in node-level tasks.

\end{itemize}

\section{Approach}
\label{sec:app}


\begin{figure*}[thpb]
\vskip 0.2in
\begin{center}
\centerline{\includegraphics[width=\textwidth]{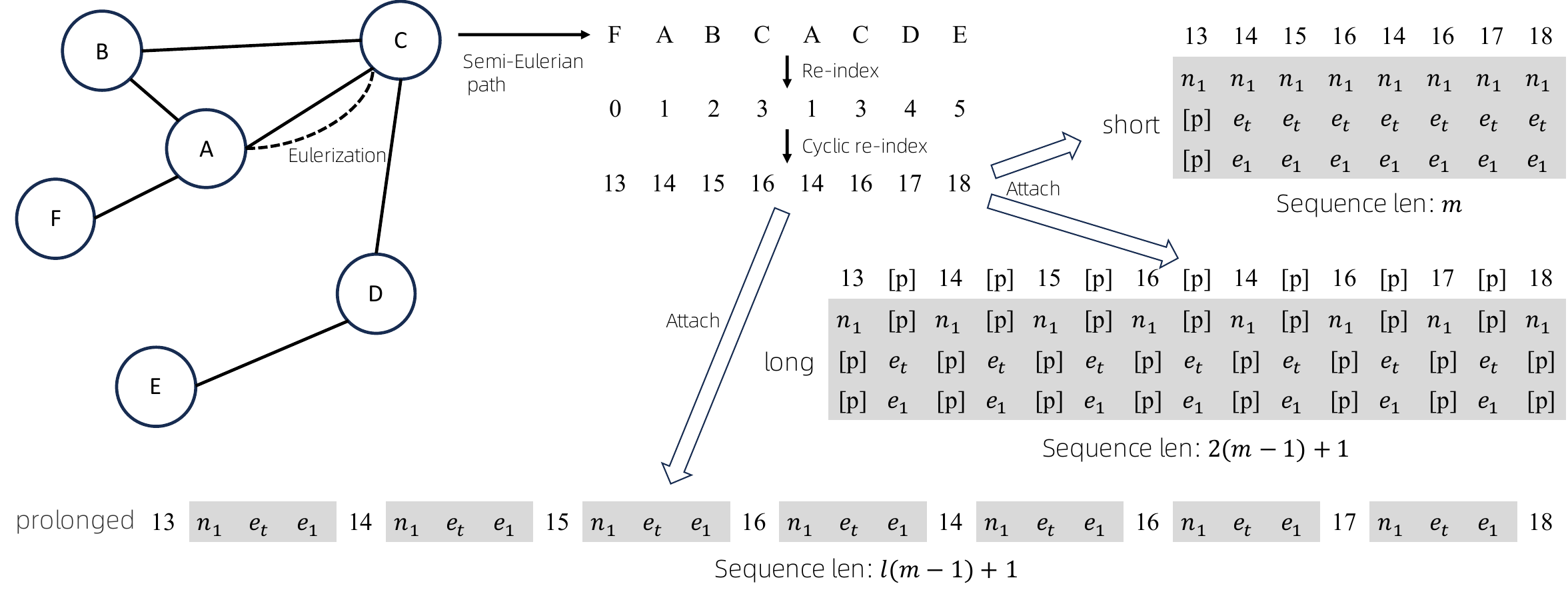}}
\caption{Overview of Graph-to-Sequence Tokenization. (Left) The process of converting a (sub)graph into a token sequence via a (semi-)Eulerian path. Dashed lines indicate duplicated edges added during Eulerization to enable full edge traversal. (Right) Three methods (short, long, prolonged) for integrating node/edge attributes into the Eulerian sequence. For simplicity, we assume one node attribute ($n_1$) and one edge attribute ($e_1$) per edge. Special tokens include padding token $\text{[p]}$ and edge type $e_t$ (e.g., incoming/outgoing direction). Sequence parameters: $m$ is length of the Eulerian sequence, and $l=2+\text{\#edge-attrs}+\text{\#node-attrs}$ (here, $l=4$).}
\label{fig:eulerize}
\end{center}
\vskip -0.2in
\end{figure*}

\subsection{Overview}

GraphGPT employs a three-stage framework: 1). \textit{Graph-to-Sequence Transformation} of GET: The input graph is converted into a sequence of tokens via (semi-)Eulerian paths, ensuring a lossless, reversible mapping between the graph and its sequential representation. This transformation preserves node, edge, and attribute information while enabling compatibility with transformer architectures. 2). \textit{Self-Supervised Pre-Training}: A standard transformer backbone (e.g., Llama; \citet{touvron2023llama}) processes these sequences using the tasks NTP or SMTP \citep{radford2018improving, chang2022maskedgit}. These tasks enable the model to learn structural and semantic graph patterns without task-specific supervision. 3). \textit{Task-Specific Fine-Tuning}:  The pre-trained model is adapted to downstream tasks—including graph classification/regression, link prediction, and node classification—by reformatting task objectives into sequence-based inputs. This unified approach maximizes the transfer of pre-trained knowledge.

\subsection{Graph to Sequence of Tokens}

To convert graphs into token sequences, we employ distinct strategies based on graph size:

\begin{itemize}

\item \textit{Small graphs} (e.g., molecular graphs) are directly serialized using the method in \S\ref{sec:serialize}.

\item \textit{Large graphs} (with up to billions of nodes/edges) are first decomposed into subgraphs via the sampling process described in \S\ref{sec:sampling}. Node identity preservation (\S\ref{sec:node-coding}) ensures structural consistency during this decomposition. These subgraphs are then serialized using \S\ref{sec:serialize}’s guidelines.

\end{itemize}

\subsubsection{Serializing Graphs with (Semi-)Eulerian Paths}
\label{sec:serialize}

We propose a \textit{lossless, reversible graph serialization} method based on traversing all edges and nodes via \textit{(semi-)Eulerian paths}. This approach guarantees:

\begin{itemize}

\item \textit{Complete representation} of nodes and edges in the sequence.

\item \textit{Bijective mapping} between the graph and its serialized form\footnote{Given a fixed starting node and predetermined node indices, the Eulerian path generated by NetworkX \citep{hagberg2008exploring} is guaranteed to be unique.}.

\end{itemize}

\textbf{Algorithmic Foundation}: The problem aligns with the Chinese Postman Problem \citep{mei1962graphic,edmonds1973matching}, which seeks the shortest path traversing all edges. For graphs lacking (semi-)Eulerian properties, we apply Eulerization \citep{edmonds1973matching,daubechies2009}, duplicating minimal edges to create an Eulerian multigraph.


\textbf{Implementation}:

1. \textit{Connectivity Check}:
    \begin{itemize}
        \item Check if the graph is connected. If not, link disconnected components by adding synthetic edges between randomly selected nodes. For example, given a graph with disconnected components A, B, and C, connect A and B via a random node pair, then B and C similarly.
        \item Label these edges with a dedicated $[$EDGE\_JUMP$]$ token and default attribute tokens.
        \item This ensures the graph becomes connected, enabling Eulerian path generation.
    \end{itemize}

2. \textit{Path Identification}:
    \begin{itemize}
        \item Check Eulerian properties using established criteria \citep{west2001introduction}.
        \item If non-Eulerian, perform Eulerization to enable path traversal.
        \item Randomly sample one valid path from possible candidates, introducing stochasticity as a data augmentation strategy akin to computer vision techniques \citep{perez2017effectiveness}. This stochasticity forces the transformer to learn invariance across different paths of the same graph and empirically reduces overfitting.
    \end{itemize}

3. \textit{Node Re-indexing}:
    \begin{itemize}
        \item Assign indices $0,1,\cdots, n-1$ based on nodes’ first appearance in the path (e.g., Fig. \ref{fig:eulerize}: node $F\rightarrow 0$, $A\rightarrow 1$).
        \item Introduce \textit{cyclic re-indexing}: $i' = (i + r) \% N$, where $r$ is a random integer and $N$ (hyperparameter) exceeds the maximum node count. 
        Without cyclic re-indexing, Eulerian paths would always start with low-index tokens (e.g., $0,1,2$), leading to skewed token frequency distributions.
        Cyclic re-indexing randomizes starting indices (e.g., selecting from 
        $\{0,1,\cdots,255\}$ for $N=256$), ensuring uniform training across all index tokens. 
        \item Cyclic re-indexing is critical for datasets like Triangles (\S \ref{sec:exp-graph-struct}), where test graphs have significantly more nodes than training graphs (e.g., test graphs up to $100$ nodes vs. training graphs $\le 25$ nodes). Without re-indexing, higher-index tokens (e.g., $25\sim 255$) remained undertrained, degrading performance.
    \end{itemize}

4. \textit{Attribute Attaching}:
    \begin{itemize}
        \item Discrete attributes: Direct tokenization.
        \item Continuous attributes: Digit-wise tokenization (e.g., $3.14 \rightarrow [3, ., 1, 4]$).
        \item Edge directionality: Distinct tokens for incoming/outgoing edges (e.g., $[\rightarrow]$, $[\leftarrow]$).
        \item Three attribute integration strategies (Fig. \ref{fig:eulerize}): \textit{short}, \textit{long}, and \textit{prolonged} formats.
    \end{itemize}

\textbf{Theoretical Guarantee:}

The serialization is \textit{lossless and reversible} (up to isomorphism) per the Eulerian Path Theorem: reconstructing edges from adjacent tokens recovers the original graph structure \citep{grohe2020graph}. For example, in Fig. \ref{fig:eulerize}’s sequence, connecting consecutive tokens ($13\rightarrow 14\rightarrow 15\rightarrow ...$) reconstructs all edges, yielding a graph isomorphic to the input.

\subsubsection{Subgraph Sampling}
\label{sec:sampling}

Directly serializing large graphs into sequences via the method in \S\ref{sec:serialize} produces excessively long sequences that exceed transformer context limits. While truncating such sequences is possible, this approach faces two critical issues:
\begin{itemize}
    \item \textit{Computational Overhead}: Eulerization and path identification for a large graph are computationally expensive.
    \item \textit{Inconsistent Training}: Sequence fragmentation introduces mismatches between pre-training and fine-tuning data formats.
\end{itemize}

To address these challenges, we adopt subgraph sampling—a scalable strategy that decomposes a large graph into smaller, manageable subgraphs before serialization.

\textbf{Implementation}:
\begin{itemize}
    \item We use the ShaDowKHop sampler \citep{zeng2021decoupling} to extract localized subgraphs centered on randomly selected nodes or edges.
    \item Sampler parameters (e.g., hop depth, neighbor count) are preconfigured to ensure generated sequences fit within the transformer’s context window. These parameters are dataset- and hardware-dependent (see App. \ref{app:subgraph} for configuration details).
\end{itemize}




\subsubsection{Node Identity Encoding}
\label{sec:node-coding}

Preserving global node identities during subgraph sampling is essential to avoid information loss. While unique token-based encoding (via learnable embeddings) is theoretically viable, it becomes impractical for graphs with billions of nodes due to:

\begin{itemize}
    \item \textit{Vocabulary Explosion}: A 10-billion-node graph would require a vocabulary of size $10^{10}$.
    \item \textit{Memory Constraints}: Corresponding embedding matrices become prohibitively large.
\end{itemize}


\textit{Solution: Multi-Token Node Encoding.} We propose encoding each node as a combination of $k$ tokens, reducing vocabulary size exponentially. For example: A $10^{10}$-node graph can be uniquely represented with two tokens from a $10^5$-size vocabulary ($10^5\times 10^5 = 10^{10}$). Graph partitioning via METIS \citep{karypis1997metis} enables this by dividing the graph into $10^5$ clusters, each containing $\sim 10^5$ nodes.

\textit{Trade-offs}: Increasing $k$ (\eg, $k=5$) allows smaller vocabularies ($100^5=10^{10}$) but lengthens sequences. This mirrors variable-length encodings like utf-8 \citep[Chapter~2]{allen2012unicode}, where characters are represented by 1–4 bytes (vocabulary size $= 256$).

Our ablation studies (\S\ref{sec:ablation-node-id}) demonstrate this method’s effectiveness in preserving node identity.




\subsection{Modeling with the Transformer Decoder/Encoder}
We demonstrate how the transformer architecture processes graph token sequences under a unified pre-training and fine-tuning paradigm for diverse graph tasks.

\subsubsection{Pre-training with the NTP or SMTP Tasks}

Self-supervised pre-training has proven critical for success in NLP  \citep{devlin2018bert,radford2019language} and CV \citep{he2022masked,chang2022maskedgit}. We adopt two foundational generative tasks: NTP, which enables SOTA performance in NLP \citep{brown2020language}, and SMTP, which extends masked prediction with scheduled masking rates.

\textbf{Implementation}:
\begin{itemize}
    \item \textit{Masking (SMTP only)}: For node-level masking, all occurrences of a masked node in the Eulerian sequence are hidden to prevent leakage (\eg, two occurrences of node $A$ in Fig. \ref{fig:eulerize} are masked concurrently).
    \item \textit{Mask Scheduling (SMTP only)}: Following \citet{chang2022maskedgit}, we adopt the same linear scheduling function, which empirically balances training stability and performance.
    \item \textit{Multi-Token Prediction (NTP and SMTP)}: For sequences encoded in \textit{short} or \textit{long} formats (Fig. \ref{fig:eulerize}), we predict all non-padding tokens per column simultaneously, similar to \citet{gloeckle2024nntp}.
\end{itemize}




\subsubsection{Fine-tuning on Downstream Graph Tasks}
\label{sec:graph-tasks}

We adapt the pre-trained transformer to supervised tasks by reformatting the sequence-based inputs, ensuring alignment with pre-training:

\begin{enumerate}
    \item \textit{Graph-Level} Tasks (e.g., classification/regression): Append a special $[$GSUM$]$ token to the sequence.
    \item \textit{Edge-Level} Tasks (e.g., link prediction): Append tokens of the target edge’s source and destination nodes.
    \item \textit{Node-Level} Tasks (e.g., node classification): Append the target node’s token to the sequence.
\end{enumerate}

The final token’s output is fed to a randomly initialized multilayer perceptron (MLP). Fig. \ref{fig:pre-train-fine-tune} (Appendix) illustrates the implementation.

The transformer weights are initialized from pre-trained checkpoints, and the MLP layers are initialized randomly. All parameters are updated during fine-tuning.


This formulation ensures seamless knowledge transfer from pre-training, mirroring successes in NLP \citep{brown2020language,wei2021finetuned}.



\section{Experiments}
\label{sec:exp}

\subsection{Datasets}
Recent advances in AI for scientific discovery \citep{wang2023ai4sci} motivate our evaluation of GraphGPT on large-scale scientific graph datasets spanning physics, chemistry, and bioinformatics. To demonstrate its versatility across graph tasks, we select benchmarks for graph-, edge-, and node-level objectives:
\begin{itemize}
    \item Graph-level: PCQM4Mv2 (quantum chemistry), ogbg-molpcba (molecular property prediction) and Triangles (triangles counting).
    \item Edge-level: ogbl-ppa (protein-protein associations) and ogbl-citation2 (citation networks).
    \item Node-level: ogbn-proteins (protein interaction networks) and ogbn-arxiv (paper categorization).
\end{itemize}

Dataset statistics are detailed in Table \ref{tab:stats} (Appendix \ref{app:data}).

\begin{itemize}
    \item PCQM4Mv2 contains $>3.7$ million organic molecules from PubChemQC \citep{nakata2017pubchemqc}. Nodes represent atoms (9D attributes: atomic number, chirality, etc.), and edges denote chemical bonds (3D attributes: bond type, stereochemistry, conjugation).
    \item ogbg-molpcba is a smaller molecular dataset \citep{wu2017molecule} with the same node/edge attributes.
    \item Triangles \citep{knyazev2019understanding} contains 45k graphs (no node/edge attributes).
    \item ogbl-ppa: Nodes are proteins from 58 species; edges represent functional associations \citep{szklarczyk2019string}.
    \item ogbl-citation2: A directed citation network with $\sim 3$ million papers (nodes) and $>30$ million edges.
    \item ogbn-proteins: Undirected, weighted graph of 132,534 proteins (nodes) with 8D edge attributes encoding association strengths.
    \item ogbn-arxiv: Citation network of 169,343 papers; tasks involve predicting 40 subject categories.
\end{itemize}

Empirical results demonstrate that SMTP pre-training achieves superior or comparable performance across all benchmarks. Unless otherwise specified, reported GraphGPT results utilize SMTP pre-training.



\subsection{Graph-Level Tasks}
\label{sec:exp-graphlevel}

\begin{table}[ht]
\caption{Results of the graph regression task on the PCQM4Mv2 dataset. The metric is mean absolute error (MAE), the smaller the better. 
86\% of the valid dataset is added to training after hyper-parameters selection. 
Superscript numbers indicate source references, while subscript numbers correspond to model variants in Table \ref{tab:models} (Appendix). 
The best results are in bold, and second-best are underlined. 
This notation convention applies to all subsequent tables.
}
\label{tab:graph-pcqm4m}
\vskip 0.1in
\begin{center}
\begin{small}
\begin{tabular}{ll|cc|l}
\toprule
 \multirow{2}{*}{} & \multirow{2}{*}{Models} &  \multicolumn{2}{c|}{MAE $\downarrow$}    & \multirow{2}{*}{Params} \\ 
                   &                         &   Valid & \multicolumn{1}{c|}{Test}       &                         \\ 

\midrule

 \multirow{4}{*}{GNN} & GCN\tsup{1}             & 0.1379 & 0.1398 & 2.0M \\
                      & GIN\tsup{2}             & 0.1195 & 0.1218 & 3.8M \\
                      & GCN\tsup{1}-VN\tsup{3}  & 0.1153 & 0.1152 & 4.9M \\
                      & GIN\tsup{2}-VN\tsup{3}  & 0.1083 & 0.1084 & 6.7M \\
 
 \midrule
 
 \multirow{5}{*}{GT}  & TokenGT\tsup{4}       & 0.0910   & 0.0919    & 48.5M                       \\ 
                      & Graphformer\tsup{5}   & 0.0864   & N/A   & 48.3M                   \\ 
                      & GPS-Deep\tsup{6}      & 0.0852   & 0.0862   & 138.1M                   \\ 
                      & GPS++ (no 3D)\tsup{7} & 0.0818   & N/A   & 40.0M                   \\ 
                      & GPTrans-L\tsup{8}     & \underline{0.0809}   & \underline{0.0821}   & 86.0M                  \\

\midrule

 \multirow{4}{*}{Ours} & GraphGPT-M          & 0.0827  & N/A      &   37.7M   \\
                       & GraphGPT-B$_{12}$   & \underline{0.0807}  & N/A      &  113.6M    \\
                       & GraphGPT-B$_{24}$   & \bftab{0.0793} & N/A      &  227.3M    \\
                       & GraphGPT-B$_{48}$   & \bftab{0.0792}    & \bftab{0.0804}   & 453.4M    \\

\bottomrule
\end{tabular}
\newline
\vskip 0.01in
\scriptsize{\tsup{1}\citet{kipf2017semi}, \tsup{2}\citet{xu2019how}, \tsup{3}\citet{gilmer2017}, \tsup{4}\citet{kim2022pure}, \tsup{5}\citet{ying2021transformers}, \tsup{6}\citet{rampasek2022recipe}, \tsup{7}\citet{masters2023gps_pp}, \tsup{8}\citet{chen2023gptrans}
}
\end{small}
\end{center}
\vskip -0.1in
\end{table}

We evaluate GraphGPT on two molecular datasets where tasks involve predicting quantum chemical properties solely from 2D molecular graphs—a practical alternative to relying on 3D equilibrium structures. Specifically, PCQM4Mv2 predicts the HOMO-LUMO energy gap, and ogbg-molpcba predicts 128 binary molecular properties.


On PCQM4Mv2, GraphGPT achieves a test MAE of \textbf{0.0804},  significantly outperforming the previous SOTA (0.0821, \citet{chen2023gptrans}).

Compared to GTs like TokenGT, Graphformer, GPS, and GPTrans—which require handcrafted features or intricate architectures to encode structural information—GraphGPT attains superior performance without manual feature engineering. It also surpasses GNNs by a substantial margin (Table \ref{tab:graph-pcqm4m}).

\textbf{Analysis} (Tables \ref{tab:graph-pcqm4m} and \ref{tab:graph-pcba}):

\textit{Lossless Serialization}: The Eulerian path-based serialization and generative pre-training enable GraphGPT to fully capture structural and semantic graph information.

\textit{Scalability}: While GTs with fewer parameters often plateau when scaled \citep{shi2022benchmark}, GraphGPT shows consistent improvement up to \textbf{200M} parameters. The log-log scaling law plot for both pre-training loss and supervised fine-tuning loss is shown in Fig.~\ref{fig:scaling-law-pt-sft} (Appendix).

\textit{Parameter Efficiency}: GraphGPT’s larger parameter count may reflect its capacity to implicitly learn features that other GTs encode manually. Generative pre-training also allocates model capacity to generation tasks, potentially limiting discriminative performance of models at smaller scales.


\textit{Limitations}: Pre-training on additional external large-scale molecular datasets yielded diminishing returns, suggesting saturation in 2D structural information. Incorporating 3D molecular data could help address this limitation.

\textit{Transfer Learning}: When fine-tuned on ogbg-molpcba, our PCQM4Mv2-pretrained model achieves results exceeding powerful GNNs (GCN, GIN) and matching SOTA GTs (Table \ref{tab:graph-pcba}).

\begin{table}[ht]
\caption{Results of the graph classification task on the ogbg-molpcba dataset. All the baseline results are from the OGB leaderboard or the corresponding papers. $^\dagger$ indicates the model is pre-trained on PCQM4M-v2 dataset.}
\label{tab:graph-pcba}
\vskip 0.1in
\begin{center}
\begin{small}
\begin{tabular}{l|cc|l}
\toprule

\multirow{2}{*}{Models}      & \multicolumn{2}{c|}{Average Precision (\%) $\uparrow$} & \multirow{2}{*}{Params} \\ 
                             & \multicolumn{1}{c}{Test}          & \multicolumn{1}{c|}{Valid}        &    \\ 

\midrule

 GCN\tsup{1}                         & $20.20_{\pm 0.24}$   & $20.59_{\pm 0.33}$   & 0.57M                   \\
 GIN\tsup{2}                         & $22.66_{\pm 0.28}$   & $23.05_{\pm 0.27}$   & 1.92M                   \\
 GINE\tsup{3}-VN\tsup{4}             & $29.17_{\pm 0.15}$   & $30.65_{\pm 0.30}$   & 6.1M                   \\
 NGIN\tsup{5}-VN\tsup{4}             & $30.07_{\pm 0.37}$   & $30.59_{\pm 0.56}$   & 44.19M                  \\
 PDF\tsup{6}                         & $30.31_{\pm 0.26}$   & $31.15_{\pm 0.20}$   & 3.84M                  \\

\midrule

 Graphormer-L$^\dagger$\tsup{7}      & $31.40_{\pm 0.32}$   & \underline{$32.27_{\pm 0.24}$}   & 119.5M                   \\
 EGT-Larger$^\dagger$\tsup{8}        & $29.61_{\pm 0.24}$   & N/A   & 110.8M                   \\
 GRPE-Large$^\dagger$\tsup{9}        & $31.50_{\pm 0.10}$   & N/A   & 118.3M                   \\
 GPTrans-L$^\dagger$\tsup{10}        & $\mathbf{32.43_{\pm 0.22}}$   & N/A   & 86.0M                   \\

\midrule

 GraphGPT-M$^\dagger$                 & $30.13_{\pm 0.25}$             &  $31.62_{\pm 0.24}$       &   37.7M               \\
 GraphGPT-B$^\dagger_{12}$            & $31.28_{\pm 0.23}$             &  \underline{$32.27_{\pm 0.15}$}  &   113.6M               \\
 GraphGPT-B$^\dagger_{24}$            & \underline{$31.81_{\pm 0.1}$}  &  $\mathbf{32.54_{\pm 0.2}}$         &   227.3M             \\

\bottomrule
\end{tabular}
\newline
\vskip 0.01in
\scriptsize{\tsup{1}\citet{kipf2017semi}, \tsup{2}\citet{xu2019how}, \tsup{3}\citet{brossard2011}, \tsup{4}\citet{gilmer2017}, \tsup{5}\citet{zhang2021nestedgnn}, \tsup{6}\citet{yang2023pdf}, \tsup{7}\citet{ying2021transformers}, \tsup{8}\citet{hussain2022egt}, \tsup{9}\citet{park2022grpe}, \tsup{10}\citet{chen2023gptrans}

}
\end{small}
\end{center}
\vskip -0.1in
\end{table}


\subsubsection{Graph Structure Understanding}
\label{sec:exp-graph-struct}

To evaluate GraphGPT’s ability to learn structural patterns through generative pre-training, we use the Triangles dataset with the task of counting triangles. The dataset is split into: 1). \textit{Training/Validation}: 30k and 5k small graphs ($\le 25$ nodes); 2). \textit{Testing}: 5k small graphs (Test-small) and 5k large graphs (25–100 nodes, Test-large).



This task is challenging even for in-distribution (ID) graphs and considerably harder for out-of-distribution (OOD) graphs.


\textbf{Pre-Training Setup}: We augment pre-training with diverse datasets, \ie, 
Reddit-threads \citep{rozemberczki2020karate}, Erd\H{o}s-R{\'e}nyi random graphs \citep{erdos1960evolution}, and Internal real-world graphs (See Table \ref{tab:stats}, Appendix \ref{app:data}).



\textbf{Analysis} (Table \ref{tab:graph-triangles}):


\emph{Pre-Training Efficacy}: GraphGPT achieves comparable accuracy to GTs on ID graphs and superior OOD generalization (lower variance). This demonstrates that generative pre-training effectively encodes structural knowledge transferable to downstream tasks.

\emph{Impact of Graph Types}: Pre-training on real-world graphs (e.g., internal datasets) outperforms random Erd\H{o}s-R{\'e}nyi graphs, suggesting meaningful structural patterns in real-world data enhance model learning.

\emph{Dataset Diversity}: Combining Triangles with diverse datasets (Reddit-threads, internal graphs) yields better performance than pre-training on Triangles alone. This highlights the importance of diverse pre-training data for learning generalizable structural patterns.

\emph{Attributed Graphs}: Models pre-trained on attributed graphs (PCQM4Mv2, ogbl-ppa, ogbn-proteins) and fine-tuned on Triangles achieve significant improvements: $64.3\%/86.1\%/86.6\%$ vs. $32.6\%$ (baseline GET without pre-training). This confirms that structural knowledge is obtained even when pre-training includes node/edge attributes.

\begin{table}[ht]
\caption{Results of the graph classification task on the Triangles dataset. Superscript numbers indicate source references, while letters denote specific pre-training datasets. We report averaged metrics from 10 independent runs to ensure statistical reliability. The baseline results are from \citet{muller2023attending}.}
\label{tab:graph-triangles}
\vskip 0.1in
\begin{center}
\begin{small}
\begin{tabular}{l|cc|l}
\toprule

\multirow{2}{*}{Models}      & \multicolumn{2}{c|}{Accuracy (\%) $\uparrow$} & \multirow{2}{*}{Params} \\ 
                             & \multicolumn{1}{c}{T-small}          & \multicolumn{1}{c|}{T-large}        &    \\ 

\midrule

 GIN\tsup{1}                         & $71.53_{\pm 0.94}$   & $33.54_{\pm 0.30}$   & 0.15M                   \\
 Transformer\tsup{2}                 & $12.08_{\pm 0.31}$   & $10.01_{\pm 0.04}$   & 0.2M                   \\

\midrule

 Transformer-LapPE\tsup{3}           & $78.29_{\pm 0.25}$   & $10.64_{\pm 2.94}$   & 0.2M                  \\
 Transformer-RWSE\tsup{3}            & $\mathbf{99.40_{\pm 0.10}}$   & \underline{$54.76_{\pm 7.24}$}   & 0.2M                  \\
 Graphormer\tsup{4}                  & $\mathbf{99.09_{\pm 0.31}}$   & $42.34_{\pm 6.48}$   & 0.2M                  \\

\midrule

 GET-B                               & $32.60_{\pm 1.86}$              &  $13.99_{\pm 1.78}$      &   113.5M               \\
 GraphGPT-B\tsup{a}                  & $92.16_{\pm 0.28}$              &  $26.51_{\pm 1.01}$       &   113.5M             \\
 GraphGPT-B\tsup{b}                  & $81.38_{\pm 0.27}$              &  $37.68_{\pm 0.99}$       &   113.5M             \\
 GraphGPT-B\tsup{c}                  & $\mathbf{99.08_{\pm 0.14}}$              &  $38.80_{\pm 3.60}$       &   113.5M             \\
 GraphGPT-B\tsup{d}                  & $90.93_{\pm 0.51}$              &  $40.79_{\pm 1.40}$       &   113.5M             \\
\midrule
 GraphGPT-B\tsup{e}                  & $64.28_{\pm 0.33}$              &  $17.38_{\pm 0.61}$       &   113.5M             \\
 GraphGPT-B\tsup{f}                  & $86.14_{\pm 7.38}$              &  $26.94_{\pm 4.80}$       &   113.5M             \\
 GraphGPT-B\tsup{g}                  & $86.57_{\pm 2.74}$              &  $23.45_{\pm 1.44}$       &   113.5M             \\
\midrule
 GraphGPT-B\tsup{a+b}                  & $84.83_{\pm 0.81}$              &  $39.62_{\pm 1.84}$       &   113.5M             \\
 GraphGPT-B\tsup{a+c}                  & \underline{$98.68_{\pm 0.18}$}              &  $50.07_{\pm 3.28}$       &   113.5M             \\
 GraphGPT-B\tsup{b+c}                  & \underline{$98.26_{\pm 0.30}$}              &  $52.33_{\pm 2.61}$       &   113.5M             \\
 \midrule
 GraphGPT-B\tsup{a+b+d}                & $89.98_{\pm 0.54}$              &  $33.45_{\pm 2.51}$       &   113.5M             \\
 GraphGPT-M\tsup{a+b+c}                & $95.07_{\pm 0.67}$              &  $51.72_{\pm 1.12}$       &   33.7M             \\
 GraphGPT-B\tsup{a+b+c}                  & \underline{$98.63_{\pm 0.18}$}              &  $\mathbf{58.96_{\pm 1.90}}$       &   113.5M             \\

\bottomrule
\end{tabular}
\newline
\vskip 0.01in
\scriptsize{\tsup{1}\citet{xu2019how}, \tsup{2}\citet{vaswani2017attention}, \tsup{3}\citet{rampasek2022recipe}, \tsup{4}\citet{ying2021transformers}
}
\vskip -0.1in
\scriptsize{
Pre-trained with: \tsup{a}Triangles (45K), \tsup{b}Reddit-threads (0.2M), \tsup{c}Internal dataset (3.1M), \tsup{d}Random graphs (3.1M),
\tsup{e}PCQM4M-v2 (3.7M), \tsup{f}ogbl-ppa (1), \tsup{g}ogbn-proteins (1).
}
\end{small}
\end{center}
\vskip -0.1in
\end{table}

\subsection{Edge-Level Tasks}
\label{sec:exp-edgelevel}

We evaluate GraphGPT on link prediction using the ogbl-ppa and ogbl-citation2  datasets. Results are summarized in Table \ref{tab:edge}.

\textit{Performance Superiority}: GraphGPT significantly outperforms all baseline methods, including GNNs, heuristic models, and latent-factor approaches, across both datasets. This underscores the effectiveness of generative pre-training and sequence-based modeling for edge-level tasks.

\textit{Scalability}: GraphGPT scales seamlessly to \textit{2 billion parameters}, achieving sustained performance gains with increasing model size. This motivates future exploration of even larger architectures and datasets.

\textit{Transformer Efficacy}: To our knowledge, GraphGPT is the first transformer-based model to achieve SOTA results on ogbl-ppa and ogbl-citation2, demonstrating the viability of sequence-driven architectures for large-scale edge-level tasks.

\begin{table}[ht]
\caption{Results of the link prediction task on the ogbl-ppa and ogbl-citation2 datasets.}
\label{tab:edge}
\vskip 0.1in
\begin{center}
\begin{small}
\begin{tabular}{l|c|c}
\toprule
 \multirow{2}{*}{Models}      & ogbl-ppa                  & ogbl-citation2 \\ 
                              & HR@100 (\%) $\uparrow$    & MRR (\%) $\uparrow$ \\ 

\midrule

 Common Neighbor              & $27.65_{\pm 0.00}$    & $51.47_{\pm 0.00}$                       \\ 
 Adamic Adar                  & $32.45_{\pm 0.00}$    & $51.89_{\pm 0.00}$                       \\ 
 Resource Allocation\tsup{1}          & $49.33_{\pm 0.00}$    & $51.98_{\pm 0.00}$                       \\ 

\midrule

 Node2Vec\tsup{2}                     & $22.26_{\pm 0.83}$    & $61.41_{\pm 0.11}$                 \\ 
 Matrix Factorization\tsup{3}         & $32.29_{\pm 0.94}$    & $51.86_{\pm 4.43}$                 \\ 

\midrule

 GCN\tsup{4}                          & $18.67_{\pm 1.32}$    & $84.74_{\pm 0.21}$                   \\ 
 GraphSAGE\tsup{5}                    & $16.55_{\pm 2.40}$    & $82.60_{\pm 0.36}$                   \\ 
 SEAL\tsup{6}                         & $48.80_{\pm 3.16}$    & $87.67_{\pm 0.32}$                   \\ 
 AGDN\tsup{7}                         & $41.23_{\pm 1.59}$    & $85.49_{\pm 0.29}$                  \\ 
 SIEG\tsup{8}                         & $63.22_{\pm 1.74}$    & $90.18_{\pm 0.15}$                   \\
 MPLP\tsup{9}                         & $65.24_{\pm 1.50}$     & $90.72_{\pm 0.12}$                  \\
 RefinedGAE\tsup{10}                   & \underline{$73.74_{\pm 0.92}$}    & $84.55_{\pm 0.15}$      \\

\midrule

 GraphGPT-M        & $65.44_{\pm 0.43}$     & \underline{$92.82_{\pm 0.27}$}      \\
 GraphGPT-B          & $68.76_{\pm 0.67}$    & $\mathbf{93.05_{\pm 0.20}}$                 \\ 
 GraphGPT-XXL       & $\mathbf{76.55_{\pm 0.67}}$    & N/A      \\ 

\bottomrule
\end{tabular}
\newline
\vskip 0.01in
\scriptsize{\tsup{1}\citet{zhou2009predicting}, \tsup{2}\citet{grover2016node2vec}, \tsup{3}\citet{mnih2008probabilistic}, \tsup{4}\citet{kipf2017semi}, \tsup{5}\citet{hamilton2017inductive}, \tsup{6}\citet{zhang2021labeling}, \tsup{7}\citet{sun2020adaptive}, \tsup{8}\citet{shi2024sieg}, \tsup{9}\citet{dong2023mplp}, \tsup{10}\citet{ma2024refinedgae}
}
\end{small}
\end{center}
\vskip -0.1in
\end{table}

\subsection{Node-Level Tasks}
\label{sec:exp-nodelevel}

We evaluate GraphGPT on two node-level benchmarks: ogbn-proteins predicts 112 binary protein function labels, and ogbn-arxiv classifies arXiv papers into 40 subject categories. Results are summarized in Table \ref{tab:node}.

\textit{ogbn-proteins}: GraphGPT surpasses well-tuned GNN baselines (GCN, GraphSAGE, GAT) and significantly outperforms graph transformers (GTs). Remarkably, GraphGPT achieves competitive performance with input subgraphs of $\sim 40$ nodes, while SOTA GNNs like AGDN \citep{sun2020adaptive} require subgraphs with $>22,000$ nodes.

\textit{ogbn-arxiv}: GraphGPT delivers performance comparable to or approaching SOTA graph transformers and optimized GNNs.

The strong performance with minimal neighborhood sampling suggests that generative pre-training effectively encodes global structural and semantic graph information into node token embeddings and transformer parameters. This contrasts with traditional GNNs, which rely on extensive local aggregation with feature propagation.





\begin{table}[ht]
\caption{Results of the node classification task on the ogbn-proteins and ogbn-arxiv datasets.}
\label{tab:node}
\vskip 0.1in
\begin{center}
\begin{small}
\begin{tabular}{l|l|l}
\toprule
\multirow{2}{*}{Models}      & ogbn-proteins               & ogbn-arxiv         \\
                             & ROC-AUC (\%) $\uparrow$     & Accuracy (\%) $\uparrow$    \\
\midrule
GCN\tsup{1,2}                & $77.29_{\pm 0.46}$             & \underline{$73.53_{\pm 0.12}$}                \\
GraphSAGE\tsup{1,3}          & $82.21_{\pm 0.32}$             & $73.00_{\pm 0.28}$                   \\
GAT\tsup{1,4}                & \underline{$85.01_{\pm 0.46}$}             & \underline{$73.30_{\pm 0.18}$}                  \\
DRGAT\tsup{5}                & N/A                            & $\mathbf{74.16_{\pm 0.07}}$          \\
AGDN\tsup{6}                 & $\mathbf{88.65_{\pm 0.13}}$    & \underline{$73.41_{\pm 0.25}$}                  \\
DeeperGCN\tsup{7}            & \underline{$85.80_{\pm 0.17}$} & $71.92_{\pm 0.16}$                   \\
\midrule
GraphGPS\tsup{1,8}           & $77.15_{\pm 0.64}$             & $71.23_{\pm 0.59}$ \\
NAGphormer\tsup{1,9}         & $72.17_{\pm 0.45}$             & $70.88_{\pm 0.24}$ \\
Exphormer\tsup{1,10}         & $77.62_{\pm 0.33}$             & $72.32_{\pm 0.36}$ \\
GOAT\tsup{1,11}              & $79.31_{\pm 0.42}$             & $72.76_{\pm 0.29}$ \\
NodeFormer\tsup{1,12}        & $77.86_{\pm 0.84}$             & $67.78_{\pm 0.28}$ \\
SGFormer\tsup{1,13}          & $79.92_{\pm 0.48}$             & $72.76_{\pm 0.33}$ \\
Polynormer\tsup{1,14}        & $79.53_{\pm 0.67}$             & \underline{$73.40_{\pm 0.22}$} \\
\midrule
GraphGPT-S                  & $83.56_{\pm 0.16}$              & $70.83_{\pm 0.33}$                  \\
GraphGPT-M                  & $84.02_{\pm 0.21}$              & $71.20_{\pm 0.34}$                  \\
GraphGPT-B                  & \underline{$85.33_{\pm 0.10}$}  & $72.10_{\pm 0.30}$                 \\
\bottomrule
\end{tabular}
\newline
\vskip 0.01in
\scriptsize{\tsup{1}\citet{luo2024classic}, \tsup{2}\citet{kipf2017semi}, \tsup{3}\citet{hamilton2017inductive}, \tsup{4}\citet{vaswani2017attention}, \tsup{5}\citet{zhang2023drgcn}, \tsup{6}\citet{sun2020adaptive}, 
\tsup{7}\citet{li2020deepergcn}, 
\tsup{8}\citet{rampasek2022recipe}, \tsup{9}\citet{chen2023nagphormer}, \tsup{10}\citet{shirzad2023exphormer}, \tsup{11}\citet{kong2023goat}, \tsup{12}\citet{wu2022nodeformer}, \tsup{13}\citet{wu2024sgformer}, \tsup{14}\citet{deng2024polynormer}
}
\end{small}
\end{center}
\vskip -0.1in
\end{table}

\subsection{Ablation Study}
\label{sec:ablation}

We analyze the impact of three core components of GraphGPT: pre-training, node re-indexing and node identity encoding.

\subsubsection{Pre-training}

The self-supervised NTP or SMTP tasks are central to GraphGPT’s success. As shown in Table \ref{tab:ablation-pretrain}, pre-training delivers performance improvements of 10–100\% across graph-, edge-, and node-level tasks. These gains highlight its role in enabling the model to learn intrinsic graph structural patterns and capture semantic relationships inherent in node and edge attributes.

\begin{table}[ht]
\caption{Ablation study of pre-training on the datasets of various types of tasks. Superscripts {\scriptsize $D/E$} stand for transformer decoder/encoder. $*$ means both molpcba and PCQM4Mv2 datasets are used for SMTP pre-training, and $\dagger$ indicates that the model is further trained using PCQM4M-v2's regression task. For the PCQM4Mv2 dataset, the metric is MAE, the lower the better.
}
\label{tab:ablation-pretrain}
\vskip 0.1in
\begin{center}
\begin{small}
\begin{sc}
\begin{tabular}{l|c|cc}
\toprule
  Datasets  & Pre-training & Test & Valid       \\
\midrule

 \multicolumn{1}{l|}{\multirow{4}{*}{PCQM4Mv2}}     &  \ding{55}$^D$  & N/A & 0.0978   \\
 & \ding{55}$^E$  & N/A & 0.0856   \\
 & NTP            & N/A & 0.0875   \\
 & SMTP           & N/A & \bftab0.0807  \\

\midrule

 \multicolumn{1}{l|}{\multirow{6}{*}{ogbg-molpcba}} &   \ding{55}$^D$  & 12.80 & 13.31\\
 & \ding{55}$^E$   & 25.80        & 26.33 \\
 & NTP             & 23.85        & 27.77 \\
 & SMTP            & 27.56        & 28.74  \\
 & SMTP$^*$        & 27.20        & 28.49  \\
 & SMTP$^*$ + FT$^\dagger$    & \bftab28.07  & \bftab29.01  \\


\midrule

  \multicolumn{1}{l|}{\multirow{4}{*}{ogbl-ppa}}    &  \ding{55}$^D$ & 41.28 &  40.14 \\
 & \ding{55}$^E$ & 42.13 &  41.57 \\
 & NTP           & 55.56 &  54.87 \\
 & SMTP          & \bftab55.68     &  \bftab54.93 \\

\midrule

 \multicolumn{1}{l|}{\multirow{4}{*}{ogbn-proteins}} &  \ding{55}$^D$ & 57.52 & 61.19 \\
 & \ding{55}$^E$ & 53.20 & 56.39 \\
 & NTP           & 75.61 & 80.47 \\
 & SMTP          & \bftab83.56        & \bftab87.73 \\

\bottomrule

\end{tabular}
\end{sc}
\end{small}
\end{center}
\vskip -0.1in
\end{table}

\subsubsection{Node Re-indexing}
\label{sec:ablation-node-reindex}

As illustrated in Fig. \ref{fig:eulerize}, we re-index the nodes based on their order in the (semi-)Eulerian path. To evaluate the effectiveness of this approach, we conduct experiments on the ogbg-molpcba dataset, with results summarized in Tab. \ref{tab:ablation-node-reindex}. 

While node re-indexing increases pre-training loss, it consistently improves performance on downstream tasks across various model sizes. This technique acts as a form of data augmentation, preventing the model from memorizing graph-specific artifacts—such as arbitrary node labeling—and thereby enhancing generalization. 

Furthermore, re-indexing enables constrained decoding of node tokens during graph generation with GraphGPT, reducing the search space for valid outputs.

\begin{table}[ht]
\caption{Ablation study of node re-indexing on the ogbg-molpcba dataset with two model sizes. PT means pre-training.}
\label{tab:ablation-node-reindex}
\vskip 0.1in
\begin{center}
\begin{small}
\begin{sc}
\begin{tabular}{l|c|c|cc}
\toprule
Params & Re-index & PT Loss & Test & Valid       \\

\midrule

\multicolumn{1}{l|}{\multirow{2}{*}{4.48M}}  &   \ding{55} & \bftab0.0844 & 23.10 & 25.25\\
                                              &   \ding{51} & 0.0874       & \bftab23.85 & \bftab27.77\\

\midrule

\multicolumn{1}{l|}{\multirow{2}{*}{114.12M}}  &  \ding{55} & \bftab0.0689 & 22.70 & 26.21\\
                                               &  \ding{51} & 0.0750       & \bftab25.17 & \bftab28.57\\

\bottomrule

\end{tabular}
\end{sc}
\end{small}
\end{center}
\vskip -0.1in
\end{table}

\subsubsection{Node Identity Encoding}
\label{sec:ablation-node-id}

Node identity encoding (see \S\ref{sec:node-coding})—representing nodes' identity in large graphs as multiple tokens—is critical for edge- and node-level tasks. Using GraphGPT-mini (a lightweight variant to conserve computational resources), we demonstrate that this method significantly enhances performance (Table \ref{tab:ablation-node-coding}). Further implementation details are provided in \cref{app:edge,app:data}.


\begin{table}[ht]
\caption{Ablation study of node identity encoding (NIE) on the ogbl-ppa and ogbn-proteins datasets.}
\label{tab:ablation-node-coding}
\vskip 0.1in
\begin{center}
\begin{small}
\begin{sc}
\begin{tabular}{l|l|c|cc}
\toprule
 Datasets  & Params & NIE & Test & Valid       \\

\midrule

 \multicolumn{1}{l|}{\multirow{2}{*}{ogbl-ppa}} &  \multicolumn{1}{l|}{\multirow{2}{*}{14.75M}}   & \ding{55} & 44.38 & 45.08 \\
                  & & \ding{51} & \bftab55.56 & \bftab54.87 \\

\midrule

   \multicolumn{1}{l|}{\multirow{2}{*}{ogbn-proteins}} &  \multicolumn{1}{l|}{\multirow{2}{*}{10.76M}}  & \ding{55} & 60.22 & 65.66 \\
                  & & \ding{51} & \bftab75.61 & \bftab80.47 \\

\bottomrule
\end{tabular}
\end{sc}
\end{small}
\end{center}
\vskip -0.1in
\end{table}

\section{Limitations}
\label{app:discussion}

We critically assess the limitations of GraphGPT to contextualize its applicability and inspire future improvements.

\textbf{\emph{Transferability.}} GraphGPT’s reliance on dataset-specific pre-training limits its ability to generalize across domains with divergent semantics (e.g., social networks vs. molecular graphs). However, it demonstrates robust \textit{cross-dataset structural understanding} (\S\ref{sec:exp-graph-struct}) and effective \textit{intra-domain transferability}, as evidenced by molecular data experiments (\S\ref{sec:exp-graphlevel}).



\textbf{\emph{Dataset size.}} Performance on some small- to medium-sized datasets (e.g., ogbn-arxiv) lags behind traditional GNNs. This can be mitigated by expanding datasets with semantically aligned data.

\textbf{\emph{Computational Cost.}} Pre-training on large-scale graphs (ogbn-proteins, ogbl-ppa) or extensive small graphs (PCQM4Mv2) with 50M+ parameters is resource-intensive. For example, pre-training GraphGPT-B (100M+ parameters) on PCQM4Mv2 with $1\times 10^{9}$ tokens requires $\sim 63$ V100 GPU hours, and fine-tuning incurs $\sim 3$ V100 GPU hours per epoch in the distributed data parallel setting with 4 GPUs. 

While GraphGPT is less practical for small datasets due to compute-performance trade-offs, it excels with large-scale data. Emerging techniques like quantization \citep{dettmers2022bit8,frantar2022gptq}, distributed training frameworks \citep{rasley2020deepspeed, shoeybi2019megatron}, and transformer optimizations \citep{dao2023flashattention2} are poised to alleviate these costs.

\textbf{\emph{Future Directions.}} These limitations highlight opportunities for research in cross-domain transfer, data-efficient training, and scalable architectures.

\section{Related Works}
\label{sec:rel-works}

\textbf{\emph{Graph Neural Networks (GNNs)}} GNNs have dominated graph learning for decades, with numerous variants achieving strong performance across tasks \citep{wu2020comprehensive}. However, they face fundamental limitations such as over-smoothing and over-squashing \citep{rusch2023survey,alon2021on}, which hinder their scalability and ability to model long-range dependencies.

\textbf{\emph{Graph Transformers (GTs)}} Inspired by transformers’ success in NLP and CV, recent work has adapted these architectures to graphs \citep{ying2021transformers,rampasek2022recipe,muller2023attending}. While GTs achieve competitive results on large-scale graph-level tasks \citep{muller2023attending}, they typically rely on handcrafted structural features or GNN-based modules to encode graph topology—either in input representations \citep{ying2021transformers,kim2022pure,masters2023gps_pp} or attention mechanisms \citep{ying2021transformers,chen2022structure,luo2022one}.

\textbf{\emph{Pre-training and fine-tuning}} The self-supervised pre-training and supervised fine-tuning paradigm, popularized by transformers \citep{vaswani2017attention}, revolutionized NLP  \citep{devlin2018bert,radford2018improving}. Scaling this approach with web-scale data \citep{brown2020language} and techniques like instruction tuning  \citep{wei2021finetuned} or reinforcement learning from human feedback \citep{ouyang2022training} further advanced the field. 
In CV, self-supervised methods like MAE \citet{he2022masked} and MaskGIT \citep{chang2022maskedgit} demonstrated that masked prediction tasks (e.g., reconstructing masked image patches) enable transformers to achieve SOTA results.

\section{Conclusions}
We introduce GraphGPT, a novel model built on the GET backbone, which achieves SOTA or near-SOTA performance across graph-, edge-, and node-level tasks on large-scale benchmarks. By unifying pretext and downstream tasks into a sequence-based paradigm, GraphGPT demonstrates strong transferability in capturing both structural graph patterns and domain-specific knowledge (e.g., molecular properties). Notably, scaling GraphGPT to billions of parameters yields consistent performance gains, highlighting its potential as a foundation for graph-centric foundation models.

Looking ahead, GraphGPT’s architecture is inherently scalable—capable of expanding to hundreds of billions of parameters—and offers promising avenues for integration or alignment with large language models (LLMs), bridging graph reasoning and textual intelligence.

\section*{Acknowledgements}



We thank our colleagues Guoshuai Wang and Tiange Xu for their insightful discussions and feedback. We appreciate the constructive feedback from the anonymous reviewers, which significantly improved the clarity of this paper. We thank our families for their unwavering support and encouragement throughout this project.

\section*{Impact Statement}


This paper presents work whose goal is to advance the field of 
Machine Learning. There are many potential societal consequences 
of our work, none which we feel must be specifically highlighted here.



\bibliography{example_paper}

\begin{thebibliography}{96}
\providecommand{\natexlab}[1]{#1}
\providecommand{\url}[1]{\texttt{#1}}
\expandafter\ifx\csname urlstyle\endcsname\relax
  \providecommand{\doi}[1]{doi: #1}\else
  \providecommand{\doi}{doi: \begingroup \urlstyle{rm}\Url}\fi

\bibitem[Allen et~al.(2012)Allen, Anderson, Becker, Cook, Davis, Edberg, Everson, Freytag, Iancu, Ishida, et~al.]{allen2012unicode}
Allen, J.~D., Anderson, D., Becker, J., Cook, R., Davis, M., Edberg, P., Everson, M., Freytag, A., Iancu, L., Ishida, R., et~al.
\newblock \emph{The unicode standard}.
\newblock Citeseer, 2012.

\bibitem[Alon \& Yahav(2021)Alon and Yahav]{alon2021on}
Alon, U. and Yahav, E.
\newblock On the bottleneck of graph neural networks and its practical implications.
\newblock In \emph{International Conference on Learning Representations}, 2021.
\newblock URL \url{https://openreview.net/forum?id=i80OPhOCVH2}.

\bibitem[Brossard et~al.(2020)Brossard, Frigo, and Dehaene]{brossard2011}
Brossard, R., Frigo, O., and Dehaene, D.
\newblock Graph convolutions that can finally model local structure.
\newblock \emph{CoRR}, abs/2011.15069, 2020.
\newblock URL \url{https://arxiv.org/abs/2011.15069}.

\bibitem[Brown et~al.(2020)Brown, Mann, Ryder, Subbiah, Kaplan, Dhariwal, Neelakantan, Shyam, Sastry, Askell, et~al.]{brown2020language}
Brown, T., Mann, B., Ryder, N., Subbiah, M., Kaplan, J.~D., Dhariwal, P., Neelakantan, A., Shyam, P., Sastry, G., Askell, A., et~al.
\newblock Language models are few-shot learners.
\newblock \emph{Advances in neural information processing systems}, 33:\penalty0 1877--1901, 2020.

\bibitem[Chang et~al.(2022)Chang, Zhang, Jiang, Liu, and Freeman]{chang2022maskedgit}
Chang, H., Zhang, H., Jiang, L., Liu, C., and Freeman, W.~T.
\newblock Maskgit: Masked generative image transformer.
\newblock In \emph{{IEEE/CVF} Conference on Computer Vision and Pattern Recognition, {CVPR} 2022, New Orleans, LA, USA, June 18-24, 2022}, pp.\  11305--11315. {IEEE}, 2022.
\newblock \doi{10.1109/CVPR52688.2022.01103}.
\newblock URL \url{https://doi.org/10.1109/CVPR52688.2022.01103}.

\bibitem[Chen et~al.(2022)Chen, O'Bray, and Borgwardt]{chen2022structure}
Chen, D., O'Bray, L., and Borgwardt, K.~M.
\newblock Structure-aware transformer for graph representation learning.
\newblock In Chaudhuri, K., Jegelka, S., Song, L., Szepesv{\'{a}}ri, C., Niu, G., and Sabato, S. (eds.), \emph{International Conference on Machine Learning, {ICML} 2022, 17-23 July 2022, Baltimore, Maryland, {USA}}, volume 162 of \emph{Proceedings of Machine Learning Research}, pp.\  3469--3489. {PMLR}, 2022.
\newblock URL \url{https://proceedings.mlr.press/v162/chen22r.html}.

\bibitem[Chen et~al.(2023{\natexlab{a}})Chen, Gao, Li, and He]{chen2023nagphormer}
Chen, J., Gao, K., Li, G., and He, K.
\newblock Nagphormer: {A} tokenized graph transformer for node classification in large graphs.
\newblock In \emph{The Eleventh International Conference on Learning Representations, {ICLR} 2023, Kigali, Rwanda, May 1-5, 2023}. OpenReview.net, 2023{\natexlab{a}}.
\newblock URL \url{https://openreview.net/forum?id=8KYeilT3Ow}.

\bibitem[Chen et~al.(2023{\natexlab{b}})Chen, Tan, Wang, Shen, Lu, Peng, Cheng, and Qi]{chen2023gptrans}
Chen, Z., Tan, H., Wang, T., Shen, T., Lu, T., Peng, Q., Cheng, C., and Qi, Y.
\newblock Graph propagation transformer for graph representation learning.
\newblock In \emph{Proceedings of the Thirty-Second International Joint Conference on Artificial Intelligence, {IJCAI} 2023, 19th-25th August 2023, Macao, SAR, China}, pp.\  3559--3567. ijcai.org, 2023{\natexlab{b}}.
\newblock \doi{10.24963/IJCAI.2023/396}.
\newblock URL \url{https://doi.org/10.24963/ijcai.2023/396}.

\bibitem[Dao(2024)]{dao2023flashattention2}
Dao, T.
\newblock Flashattention-2: Faster attention with better parallelism and work partitioning.
\newblock In \emph{The Twelfth International Conference on Learning Representations}, 2024.
\newblock URL \url{https://openreview.net/forum?id=mZn2Xyh9Ec}.

\bibitem[Daubechies \& Hughes(2009)Daubechies and Hughes]{daubechies2009}
Daubechies, I. and Hughes, S.
\newblock Graph theory, 2009.
\newblock URL \url{http://web.math.princeton.edu/math_alive/5/Notes1.pdf}.

\bibitem[Deng et~al.(2024)Deng, Yue, and Zhang]{deng2024polynormer}
Deng, C., Yue, Z., and Zhang, Z.
\newblock Polynormer: Polynomial-expressive graph transformer in linear time.
\newblock In \emph{The Twelfth International Conference on Learning Representations, {ICLR} 2024, Vienna, Austria, May 7-11, 2024}. OpenReview.net, 2024.
\newblock URL \url{https://openreview.net/forum?id=hmv1LpNfXa}.

\bibitem[Deng et~al.(2009)Deng, Dong, Socher, Li, Li, and Fei{-}Fei]{deng2009imagenet}
Deng, J., Dong, W., Socher, R., Li, L., Li, K., and Fei{-}Fei, L.
\newblock Imagenet: {A} large-scale hierarchical image database.
\newblock In \emph{2009 {IEEE} Computer Society Conference on Computer Vision and Pattern Recognition {(CVPR} 2009), 20-25 June 2009, Miami, Florida, {USA}}, pp.\  248--255. {IEEE} Computer Society, 2009.
\newblock \doi{10.1109/CVPR.2009.5206848}.
\newblock URL \url{https://doi.org/10.1109/CVPR.2009.5206848}.

\bibitem[Dettmers et~al.(2022)Dettmers, Lewis, Shleifer, and Zettlemoyer]{dettmers2022bit8}
Dettmers, T., Lewis, M., Shleifer, S., and Zettlemoyer, L.
\newblock 8-bit optimizers via block-wise quantization.
\newblock In \emph{The Tenth International Conference on Learning Representations, {ICLR} 2022, Virtual Event, April 25-29, 2022}. OpenReview.net, 2022.
\newblock URL \url{https://openreview.net/forum?id=shpkpVXzo3h}.

\bibitem[Devlin et~al.(2019)Devlin, Chang, Lee, and Toutanova]{devlin2018bert}
Devlin, J., Chang, M., Lee, K., and Toutanova, K.
\newblock {BERT:} pre-training of deep bidirectional transformers for language understanding.
\newblock In Burstein, J., Doran, C., and Solorio, T. (eds.), \emph{Proceedings of the 2019 Conference of the North American Chapter of the Association for Computational Linguistics: Human Language Technologies, {NAACL-HLT} 2019, Minneapolis, MN, USA, June 2-7, 2019, Volume 1 (Long and Short Papers)}, pp.\  4171--4186. Association for Computational Linguistics, 2019.
\newblock \doi{10.18653/v1/n19-1423}.
\newblock URL \url{https://doi.org/10.18653/v1/n19-1423}.

\bibitem[Dong et~al.(2024)Dong, Guo, and Chawla]{dong2023mplp}
Dong, K., Guo, Z., and Chawla, N.~V.
\newblock Pure message passing can estimate common neighbor for link prediction.
\newblock In Globersons, A., Mackey, L., Belgrave, D., Fan, A., Paquet, U., Tomczak, J.~M., and Zhang, C. (eds.), \emph{Advances in Neural Information Processing Systems 38: Annual Conference on Neural Information Processing Systems 2024, NeurIPS 2024, Vancouver, BC, Canada, December 10 - 15, 2024}, 2024.

\bibitem[Dosovitskiy et~al.(2021)Dosovitskiy, Beyer, Kolesnikov, Weissenborn, Zhai, Unterthiner, Dehghani, Minderer, Heigold, Gelly, Uszkoreit, and Houlsby]{dosovitskiy2021image}
Dosovitskiy, A., Beyer, L., Kolesnikov, A., Weissenborn, D., Zhai, X., Unterthiner, T., Dehghani, M., Minderer, M., Heigold, G., Gelly, S., Uszkoreit, J., and Houlsby, N.
\newblock An image is worth 16x16 words: Transformers for image recognition at scale.
\newblock In \emph{9th International Conference on Learning Representations, {ICLR} 2021, Virtual Event, Austria, May 3-7, 2021}. OpenReview.net, 2021.
\newblock URL \url{https://openreview.net/forum?id=YicbFdNTTy}.

\bibitem[Edmonds \& Johnson(1973)Edmonds and Johnson]{edmonds1973matching}
Edmonds, J. and Johnson, E.~L.
\newblock Matching, euler tours and the chinese postman.
\newblock \emph{Mathematical programming}, 5:\penalty0 88--124, 1973.

\bibitem[Erdos et~al.(1960)Erdos, R{\'e}nyi, et~al.]{erdos1960evolution}
Erdos, P., R{\'e}nyi, A., et~al.
\newblock On the evolution of random graphs.
\newblock \emph{Publ. math. inst. hung. acad. sci}, 5\penalty0 (1):\penalty0 17--60, 1960.

\bibitem[Fey \& Lenssen(2019)Fey and Lenssen]{fey2019}
Fey, M. and Lenssen, J.~E.
\newblock Fast graph representation learning with {PyTorch Geometric}.
\newblock In \emph{ICLR Workshop on Representation Learning on Graphs and Manifolds}, 2019.

\bibitem[Frantar et~al.(2022)Frantar, Ashkboos, Hoefler, and Alistarh]{frantar2022gptq}
Frantar, E., Ashkboos, S., Hoefler, T., and Alistarh, D.
\newblock {GPTQ:} accurate post-training quantization for generative pre-trained transformers.
\newblock \emph{CoRR}, abs/2210.17323, 2022.
\newblock \doi{10.48550/ARXIV.2210.17323}.
\newblock URL \url{https://doi.org/10.48550/arXiv.2210.17323}.

\bibitem[Gilmer et~al.(2017)Gilmer, Schoenholz, Riley, Vinyals, and Dahl]{gilmer2017}
Gilmer, J., Schoenholz, S.~S., Riley, P.~F., Vinyals, O., and Dahl, G.~E.
\newblock Neural message passing for quantum chemistry.
\newblock In Precup, D. and Teh, Y.~W. (eds.), \emph{Proceedings of the 34th International Conference on Machine Learning, {ICML} 2017, Sydney, NSW, Australia, 6-11 August 2017}, volume~70 of \emph{Proceedings of Machine Learning Research}, pp.\  1263--1272. {PMLR}, 2017.
\newblock URL \url{http://proceedings.mlr.press/v70/gilmer17a.html}.

\bibitem[Gloeckle et~al.(2024)Gloeckle, Idrissi, Rozi{\`{e}}re, Lopez{-}Paz, and Synnaeve]{gloeckle2024nntp}
Gloeckle, F., Idrissi, B.~Y., Rozi{\`{e}}re, B., Lopez{-}Paz, D., and Synnaeve, G.
\newblock Better {\&} faster large language models via multi-token prediction.
\newblock In \emph{Forty-first International Conference on Machine Learning, {ICML} 2024, Vienna, Austria, July 21-27, 2024}. OpenReview.net, 2024.
\newblock URL \url{https://openreview.net/forum?id=pEWAcejiU2}.

\bibitem[Grohe \& Schweitzer(2020)Grohe and Schweitzer]{grohe2020graph}
Grohe, M. and Schweitzer, P.
\newblock The graph isomorphism problem.
\newblock \emph{Communications of the ACM}, 63\penalty0 (11):\penalty0 128--134, 2020.

\bibitem[Grover \& Leskovec(2016)Grover and Leskovec]{grover2016node2vec}
Grover, A. and Leskovec, J.
\newblock node2vec: Scalable feature learning for networks.
\newblock In Krishnapuram, B., Shah, M., Smola, A.~J., Aggarwal, C.~C., Shen, D., and Rastogi, R. (eds.), \emph{Proceedings of the 22nd {ACM} {SIGKDD} International Conference on Knowledge Discovery and Data Mining, San Francisco, CA, USA, August 13-17, 2016}, pp.\  855--864. {ACM}, 2016.
\newblock \doi{10.1145/2939672.2939754}.
\newblock URL \url{https://doi.org/10.1145/2939672.2939754}.

\bibitem[Guo et~al.(2021)Guo, Sablayrolles, J{\'{e}}gou, and Kiela]{guo2021adversarial}
Guo, C., Sablayrolles, A., J{\'{e}}gou, H., and Kiela, D.
\newblock Gradient-based adversarial attacks against text transformers.
\newblock In Moens, M., Huang, X., Specia, L., and Yih, S.~W. (eds.), \emph{Proceedings of the 2021 Conference on Empirical Methods in Natural Language Processing, {EMNLP} 2021, Virtual Event / Punta Cana, Dominican Republic, 7-11 November, 2021}, pp.\  5747--5757. Association for Computational Linguistics, 2021.
\newblock \doi{10.18653/V1/2021.EMNLP-MAIN.464}.
\newblock URL \url{https://doi.org/10.18653/v1/2021.emnlp-main.464}.

\bibitem[Hagberg et~al.(2008)Hagberg, Swart, and S~Chult]{hagberg2008exploring}
Hagberg, A., Swart, P., and S~Chult, D.
\newblock Exploring network structure, dynamics, and function using networkx.
\newblock Technical report, Los Alamos National Lab.(LANL), Los Alamos, NM (United States), 2008.

\bibitem[Hamilton et~al.(2017)Hamilton, Ying, and Leskovec]{hamilton2017inductive}
Hamilton, W.~L., Ying, Z., and Leskovec, J.
\newblock Inductive representation learning on large graphs.
\newblock In Guyon, I., von Luxburg, U., Bengio, S., Wallach, H.~M., Fergus, R., Vishwanathan, S. V.~N., and Garnett, R. (eds.), \emph{Advances in Neural Information Processing Systems 30: Annual Conference on Neural Information Processing Systems 2017, December 4-9, 2017, Long Beach, CA, {USA}}, pp.\  1024--1034, 2017.

\bibitem[He et~al.(2022)He, Chen, Xie, Li, Doll{\'a}r, and Girshick]{he2022masked}
He, K., Chen, X., Xie, S., Li, Y., Doll{\'a}r, P., and Girshick, R.
\newblock Masked autoencoders are scalable vision learners.
\newblock In \emph{Proceedings of the IEEE/CVF conference on computer vision and pattern recognition}, pp.\  16000--16009, 2022.

\bibitem[Hoffmann et~al.(2022)Hoffmann, Borgeaud, Mensch, Buchatskaya, Cai, Rutherford, de~Las~Casas, Hendricks, Welbl, Clark, Hennigan, Noland, Millican, van~den Driessche, Damoc, Guy, Osindero, Simonyan, Elsen, Rae, Vinyals, and Sifre]{hoffmann2022training}
Hoffmann, J., Borgeaud, S., Mensch, A., Buchatskaya, E., Cai, T., Rutherford, E., de~Las~Casas, D., Hendricks, L.~A., Welbl, J., Clark, A., Hennigan, T., Noland, E., Millican, K., van~den Driessche, G., Damoc, B., Guy, A., Osindero, S., Simonyan, K., Elsen, E., Rae, J.~W., Vinyals, O., and Sifre, L.
\newblock Training compute-optimal large language models.
\newblock \emph{CoRR}, abs/2203.15556, 2022.
\newblock \doi{10.48550/ARXIV.2203.15556}.
\newblock URL \url{https://doi.org/10.48550/arXiv.2203.15556}.

\bibitem[Hou et~al.(2022)Hou, Liu, Cen, Dong, Yang, Wang, and Tang]{hou2022graphmae}
Hou, Z., Liu, X., Cen, Y., Dong, Y., Yang, H., Wang, C., and Tang, J.
\newblock Graphmae: Self-supervised masked graph autoencoders.
\newblock In Zhang, A. and Rangwala, H. (eds.), \emph{{KDD} '22: The 28th {ACM} {SIGKDD} Conference on Knowledge Discovery and Data Mining, Washington, DC, USA, August 14 - 18, 2022}, pp.\  594--604. {ACM}, 2022.
\newblock \doi{10.1145/3534678.3539321}.
\newblock URL \url{https://doi.org/10.1145/3534678.3539321}.

\bibitem[Hu et~al.(2021)Hu, Fey, Ren, Nakata, Dong, and Leskovec]{hu2021ogb}
Hu, W., Fey, M., Ren, H., Nakata, M., Dong, Y., and Leskovec, J.
\newblock Ogb-lsc: A large-scale challenge for machine learning on graphs.
\newblock \emph{arXiv preprint arXiv:2103.09430}, 2021.

\bibitem[Hussain et~al.(2022)Hussain, Zaki, and Subramanian]{hussain2022egt}
Hussain, M.~S., Zaki, M.~J., and Subramanian, D.
\newblock Global self-attention as a replacement for graph convolution.
\newblock In Zhang, A. and Rangwala, H. (eds.), \emph{{KDD} '22: The 28th {ACM} {SIGKDD} Conference on Knowledge Discovery and Data Mining, Washington, DC, USA, August 14 - 18, 2022}, pp.\  655--665. {ACM}, 2022.
\newblock \doi{10.1145/3534678.3539296}.
\newblock URL \url{https://doi.org/10.1145/3534678.3539296}.

\bibitem[Jin et~al.(2020)Jin, Li, Xu, Wang, Ji, Aggarwal, and Tang]{jin2020adversarial}
Jin, W., Li, Y., Xu, H., Wang, Y., Ji, S., Aggarwal, C., and Tang, J.
\newblock Adversarial attacks and defenses on graphs.
\newblock \emph{{SIGKDD} Explor.}, 22\penalty0 (2):\penalty0 19--34, 2020.
\newblock \doi{10.1145/3447556.3447566}.
\newblock URL \url{https://doi.org/10.1145/3447556.3447566}.

\bibitem[Kaplan et~al.(2020)Kaplan, McCandlish, Henighan, Brown, Chess, Child, Gray, Radford, Wu, and Amodei]{kaplan2020scaling}
Kaplan, J., McCandlish, S., Henighan, T., Brown, T.~B., Chess, B., Child, R., Gray, S., Radford, A., Wu, J., and Amodei, D.
\newblock Scaling laws for neural language models.
\newblock \emph{arXiv preprint arXiv:2001.08361}, 2020.

\bibitem[Karypis \& Kumar(1997)Karypis and Kumar]{karypis1997metis}
Karypis, G. and Kumar, V.
\newblock Metis: A software package for partitioning unstructured graphs, partitioning meshes, and computing fill-reducing orderings of sparse matrices.
\newblock Technical report, University of Minnesota Twin Cities, Department of Computer Science and Engineering, 1997.

\bibitem[Kim et~al.(2022)Kim, Nguyen, Min, Cho, Lee, Lee, and Hong]{kim2022pure}
Kim, J., Nguyen, D., Min, S., Cho, S., Lee, M., Lee, H., and Hong, S.
\newblock Pure transformers are powerful graph learners.
\newblock \emph{Advances in Neural Information Processing Systems}, 35:\penalty0 14582--14595, 2022.

\bibitem[Kipf \& Welling(2017)Kipf and Welling]{kipf2017semi}
Kipf, T.~N. and Welling, M.
\newblock Semi-supervised classification with graph convolutional networks.
\newblock In \emph{5th International Conference on Learning Representations, {ICLR} 2017, Toulon, France, April 24-26, 2017, Conference Track Proceedings}. OpenReview.net, 2017.
\newblock URL \url{https://openreview.net/forum?id=SJU4ayYgl}.

\bibitem[Knyazev et~al.(2019)Knyazev, Taylor, and Amer]{knyazev2019understanding}
Knyazev, B., Taylor, G.~W., and Amer, M.~R.
\newblock Understanding attention and generalization in graph neural networks.
\newblock In Wallach, H.~M., Larochelle, H., Beygelzimer, A., d'Alch{\'{e}}{-}Buc, F., Fox, E.~B., and Garnett, R. (eds.), \emph{Advances in Neural Information Processing Systems 32: Annual Conference on Neural Information Processing Systems 2019, NeurIPS 2019, December 8-14, 2019, Vancouver, BC, Canada}, pp.\  4204--4214, 2019.

\bibitem[Kong et~al.(2023)Kong, Chen, Kirchenbauer, Ni, Bruss, and Goldstein]{kong2023goat}
Kong, K., Chen, J., Kirchenbauer, J., Ni, R., Bruss, C.~B., and Goldstein, T.
\newblock Goat: A global transformer on large-scale graphs.
\newblock In \emph{International Conference on Machine Learning}, pp.\  17375--17390. PMLR, 2023.

\bibitem[Krizhevsky et~al.(2012)Krizhevsky, Sutskever, and Hinton]{alexnet2012}
Krizhevsky, A., Sutskever, I., and Hinton, G.~E.
\newblock Imagenet classification with deep convolutional neural networks.
\newblock In Pereira, F., Burges, C. J.~C., Bottou, L., and Weinberger, K.~Q. (eds.), \emph{Advances in Neural Information Processing Systems 25}, pp.\  1097--1105. Curran Associates, Inc., 2012.

\bibitem[Li et~al.(2023)Li, Xiong, Qian, Thabet, and Ghanem]{li2020deepergcn}
Li, G., Xiong, C., Qian, G., Thabet, A.~K., and Ghanem, B.
\newblock Deepergcn: Training deeper gcns with generalized aggregation functions.
\newblock \emph{{IEEE} Trans. Pattern Anal. Mach. Intell.}, 45\penalty0 (11):\penalty0 13024--13034, 2023.
\newblock \doi{10.1109/TPAMI.2023.3306930}.
\newblock URL \url{https://doi.org/10.1109/TPAMI.2023.3306930}.

\bibitem[Liu et~al.(2021)Liu, Lin, Cao, Hu, Wei, Zhang, Lin, and Guo]{liu2021swin}
Liu, Z., Lin, Y., Cao, Y., Hu, H., Wei, Y., Zhang, Z., Lin, S., and Guo, B.
\newblock Swin transformer: Hierarchical vision transformer using shifted windows.
\newblock In \emph{2021 {IEEE/CVF} International Conference on Computer Vision, {ICCV} 2021, Montreal, QC, Canada, October 10-17, 2021}, pp.\  9992--10002. {IEEE}, 2021.
\newblock \doi{10.1109/ICCV48922.2021.00986}.
\newblock URL \url{https://doi.org/10.1109/ICCV48922.2021.00986}.

\bibitem[Loshchilov \& Hutter(2019)Loshchilov and Hutter]{loshchilov2017decoupled}
Loshchilov, I. and Hutter, F.
\newblock Decoupled weight decay regularization.
\newblock In \emph{7th International Conference on Learning Representations, {ICLR} 2019, New Orleans, LA, USA, May 6-9, 2019}. OpenReview.net, 2019.
\newblock URL \url{https://openreview.net/forum?id=Bkg6RiCqY7}.

\bibitem[Luo et~al.(2023)Luo, Chen, Xu, Zheng, Liu, Wang, and He]{luo2022one}
Luo, S., Chen, T., Xu, Y., Zheng, S., Liu, T.-Y., Wang, L., and He, D.
\newblock One transformer can understand both 2d \& 3d molecular data.
\newblock In \emph{International Conference on Learning Representations}, 2023.

\bibitem[Luo et~al.(2024)Luo, Shi, and Wu]{luo2024classic}
Luo, Y., Shi, L., and Wu, X.
\newblock Classic gnns are strong baselines: Reassessing gnns for node classification.
\newblock In Globersons, A., Mackey, L., Belgrave, D., Fan, A., Paquet, U., Tomczak, J.~M., and Zhang, C. (eds.), \emph{Advances in Neural Information Processing Systems 38: Annual Conference on Neural Information Processing Systems 2024, NeurIPS 2024, Vancouver, BC, Canada, December 10 - 15, 2024}, 2024.

\bibitem[Ma et~al.(2024)Ma, Wang, Wang, and Zhang]{ma2024refinedgae}
Ma, W., Wang, Y., Wang, X., and Zhang, M.
\newblock Reconsidering the performance of {GAE} in link prediction.
\newblock \emph{CoRR}, abs/2411.03845, 2024.
\newblock \doi{10.48550/ARXIV.2411.03845}.
\newblock URL \url{https://doi.org/10.48550/arXiv.2411.03845}.

\bibitem[Masters et~al.(2023)Masters, Dean, Kl{\"{a}}ser, Li, Maddrell{-}Mander, Sanders, Helal, Beker, Fitzgibbon, Huang, Ramp{\'{a}}sek, and Beaini]{masters2023gps_pp}
Masters, D., Dean, J., Kl{\"{a}}ser, K., Li, Z., Maddrell{-}Mander, S., Sanders, A., Helal, H., Beker, D., Fitzgibbon, A.~W., Huang, S., Ramp{\'{a}}sek, L., and Beaini, D.
\newblock {GPS++:} reviving the art of message passing for molecular property prediction.
\newblock \emph{Trans. Mach. Learn. Res.}, 2023, 2023.
\newblock URL \url{https://openreview.net/forum?id=moVEUgJaHO}.

\bibitem[Mei-Ko(1962)]{mei1962graphic}
Mei-Ko, K.
\newblock Graphic programming using odd or even points.
\newblock \emph{Chinese Math}, 1:\penalty0 237--277, 1962.

\bibitem[Midjourney(2023)]{midjourney2023}
Midjourney, I.
\newblock Midjourney, 2023.
\newblock URL \url{https://www.midjourney.com}.

\bibitem[Min et~al.(2022)Min, Chen, Bian, Xu, Zhao, Huang, Zhao, Huang, Ananiadou, and Rong]{min2022transformer}
Min, E., Chen, R., Bian, Y., Xu, T., Zhao, K., Huang, W., Zhao, P., Huang, J., Ananiadou, S., and Rong, Y.
\newblock Transformer for graphs: An overview from architecture perspective.
\newblock \emph{CoRR}, abs/2202.08455, 2022.
\newblock URL \url{https://arxiv.org/abs/2202.08455}.

\bibitem[Mnih \& Salakhutdinov(2008)Mnih and Salakhutdinov]{mnih2008probabilistic}
Mnih, A. and Salakhutdinov, R.~R.
\newblock Probabilistic matrix factorization.
\newblock In \emph{Advances in neural information processing systems}, pp.\  1257--1264, 2008.

\bibitem[M{\"{u}}ller et~al.(2024)M{\"{u}}ller, Galkin, Morris, and Ramp{\'{a}}sek]{muller2023attending}
M{\"{u}}ller, L., Galkin, M., Morris, C., and Ramp{\'{a}}sek, L.
\newblock Attending to graph transformers.
\newblock \emph{Trans. Mach. Learn. Res.}, 2024, 2024.
\newblock URL \url{https://openreview.net/forum?id=HhbqHBBrfZ}.

\bibitem[Nakata \& Shimazaki(2017)Nakata and Shimazaki]{nakata2017pubchemqc}
Nakata, M. and Shimazaki, T.
\newblock Pubchemqc project: {A} large-scale first-principles electronic structure database for data-driven chemistry.
\newblock \emph{J. Chem. Inf. Model.}, 57\penalty0 (6):\penalty0 1300--1308, 2017.
\newblock \doi{10.1021/acs.jcim.7b00083}.
\newblock URL \url{https://doi.org/10.1021/acs.jcim.7b00083}.

\bibitem[{OGB Team}({2020})]{ogbl-ppa-web}
{OGB Team}.
\newblock {Dataset ogbl-ppa}.
\newblock \url{https://ogb.stanford.edu/docs/linkprop/#ogbl-ppa}, {2020}.
\newblock {Accessed: 2025-05-20;}.

\bibitem[Open-AI(2023)]{openai2022chatgpt}
Open-AI.
\newblock Chatgpt.
\newblock \url{https://chat.openai.com/}, 2023.

\bibitem[Ouyang et~al.(2022)Ouyang, Wu, Jiang, Almeida, Wainwright, Mishkin, Zhang, Agarwal, Slama, Ray, et~al.]{ouyang2022training}
Ouyang, L., Wu, J., Jiang, X., Almeida, D., Wainwright, C., Mishkin, P., Zhang, C., Agarwal, S., Slama, K., Ray, A., et~al.
\newblock Training language models to follow instructions with human feedback.
\newblock \emph{Advances in Neural Information Processing Systems}, 35:\penalty0 27730--27744, 2022.

\bibitem[Park et~al.(2022)Park, Chang, Lee, Kim, and seung-won hwang]{park2022grpe}
Park, W., Chang, W.-G., Lee, D., Kim, J., and seung-won hwang.
\newblock {GRPE}: Relative positional encoding for graph transformer.
\newblock In \emph{ICLR2022 Machine Learning for Drug Discovery}, 2022.
\newblock URL \url{https://openreview.net/forum?id=GNfAFN_p1d}.

\bibitem[Perez \& Wang(2017)Perez and Wang]{perez2017effectiveness}
Perez, L. and Wang, J.
\newblock The effectiveness of data augmentation in image classification using deep learning.
\newblock \emph{arXiv preprint arXiv:1712.04621}, 2017.

\bibitem[Qiu et~al.(2020)Qiu, Chen, Dong, Zhang, Yang, Ding, Wang, and Tang]{qiu2020gcc}
Qiu, J., Chen, Q., Dong, Y., Zhang, J., Yang, H., Ding, M., Wang, K., and Tang, J.
\newblock {GCC:} graph contrastive coding for graph neural network pre-training.
\newblock In Gupta, R., Liu, Y., Tang, J., and Prakash, B.~A. (eds.), \emph{{KDD} '20: The 26th {ACM} {SIGKDD} Conference on Knowledge Discovery and Data Mining, Virtual Event, CA, USA, August 23-27, 2020}, pp.\  1150--1160. {ACM}, 2020.
\newblock \doi{10.1145/3394486.3403168}.
\newblock URL \url{https://doi.org/10.1145/3394486.3403168}.

\bibitem[Radford et~al.(2018)Radford, Narasimhan, Salimans, and Sutskever]{radford2018improving}
Radford, A., Narasimhan, K., Salimans, T., and Sutskever, I.
\newblock Improving language understanding by generative pre-training, 2018.

\bibitem[Radford et~al.(2019)Radford, Wu, Child, Luan, Amodei, Sutskever, et~al.]{radford2019language}
Radford, A., Wu, J., Child, R., Luan, D., Amodei, D., Sutskever, I., et~al.
\newblock Language models are unsupervised multitask learners.
\newblock \emph{OpenAI blog}, 1\penalty0 (8):\penalty0 9, 2019.

\bibitem[Raffel et~al.(2020)Raffel, Shazeer, Roberts, Lee, Narang, Matena, Zhou, Li, and Liu]{raffel2020exploring}
Raffel, C., Shazeer, N., Roberts, A., Lee, K., Narang, S., Matena, M., Zhou, Y., Li, W., and Liu, P.~J.
\newblock Exploring the limits of transfer learning with a unified text-to-text transformer.
\newblock \emph{The Journal of Machine Learning Research}, 21\penalty0 (1):\penalty0 5485--5551, 2020.

\bibitem[Ramp{\'{a}}sek et~al.(2022)Ramp{\'{a}}sek, Galkin, Dwivedi, Luu, Wolf, and Beaini]{rampasek2022recipe}
Ramp{\'{a}}sek, L., Galkin, M., Dwivedi, V.~P., Luu, A.~T., Wolf, G., and Beaini, D.
\newblock Recipe for a general, powerful, scalable graph transformer.
\newblock In \emph{NeurIPS}, 2022.

\bibitem[Rasley et~al.(2020)Rasley, Rajbhandari, Ruwase, and He]{rasley2020deepspeed}
Rasley, J., Rajbhandari, S., Ruwase, O., and He, Y.
\newblock Deepspeed: System optimizations enable training deep learning models with over 100 billion parameters.
\newblock In \emph{Proceedings of the 26th ACM SIGKDD International Conference on Knowledge Discovery \& Data Mining}, pp.\  3505--3506, 2020.

\bibitem[Rozemberczki et~al.(2020)Rozemberczki, Kiss, and Sarkar]{rozemberczki2020karate}
Rozemberczki, B., Kiss, O., and Sarkar, R.
\newblock Karate club: An {API} oriented open-source python framework for unsupervised learning on graphs.
\newblock In d'Aquin, M., Dietze, S., Hauff, C., Curry, E., and Cudr{\'{e}}{-}Mauroux, P. (eds.), \emph{{CIKM} '20: The 29th {ACM} International Conference on Information and Knowledge Management, Virtual Event, Ireland, October 19-23, 2020}, pp.\  3125--3132. {ACM}, 2020.
\newblock \doi{10.1145/3340531.3412757}.
\newblock URL \url{https://doi.org/10.1145/3340531.3412757}.

\bibitem[Rusch et~al.(2023)Rusch, Bronstein, and Mishra]{rusch2023survey}
Rusch, T.~K., Bronstein, M.~M., and Mishra, S.
\newblock A survey on oversmoothing in graph neural networks.
\newblock \emph{CoRR}, abs/2303.10993, 2023.
\newblock \doi{10.48550/ARXIV.2303.10993}.
\newblock URL \url{https://doi.org/10.48550/arXiv.2303.10993}.

\bibitem[Sennrich et~al.(2016)Sennrich, Haddow, and Birch]{sennrich2016neural}
Sennrich, R., Haddow, B., and Birch, A.
\newblock Neural machine translation of rare words with subword units.
\newblock In \emph{Proceedings of the 54th Annual Meeting of the Association for Computational Linguistics, {ACL} 2016, August 7-12, 2016, Berlin, Germany, Volume 1: Long Papers}. The Association for Computer Linguistics, 2016.
\newblock \doi{10.18653/v1/p16-1162}.
\newblock URL \url{https://doi.org/10.18653/v1/p16-1162}.

\bibitem[Shao et~al.(2022)Shao, Shi, Yi, Chen, and Hsieh]{shao2022adversarial}
Shao, R., Shi, Z., Yi, J., Chen, P., and Hsieh, C.
\newblock On the adversarial robustness of vision transformers.
\newblock \emph{Trans. Mach. Learn. Res.}, 2022, 2022.
\newblock URL \url{https://openreview.net/forum?id=lE7K4n1Esk}.

\bibitem[Shi et~al.(2024)Shi, Hu, Zhao, He, Zhang, and Zhou]{shi2024sieg}
Shi, L., Hu, B., Zhao, D., He, J., Zhang, Z., and Zhou, J.
\newblock Structural information enhanced graph representation for link prediction.
\newblock In Wooldridge, M.~J., Dy, J.~G., and Natarajan, S. (eds.), \emph{Thirty-Eighth {AAAI} Conference on Artificial Intelligence, {AAAI} 2024, Thirty-Sixth Conference on Innovative Applications of Artificial Intelligence, {IAAI} 2024, Fourteenth Symposium on Educational Advances in Artificial Intelligence, {EAAI} 2014, February 20-27, 2024, Vancouver, Canada}, pp.\  14964--14972. {AAAI} Press, 2024.
\newblock \doi{10.1609/AAAI.V38I13.29417}.
\newblock URL \url{https://doi.org/10.1609/aaai.v38i13.29417}.

\bibitem[Shi et~al.(2022)Shi, Zheng, Ke, Shen, You, He, Luo, Liu, He, and Liu]{shi2022benchmark}
Shi, Y., Zheng, S., Ke, G., Shen, Y., You, J., He, J., Luo, S., Liu, C., He, D., and Liu, T.
\newblock Benchmarking graphormer on large-scale molecular modeling datasets.
\newblock \emph{CoRR}, abs/2203.04810, 2022.
\newblock \doi{10.48550/ARXIV.2203.04810}.
\newblock URL \url{https://doi.org/10.48550/arXiv.2203.04810}.

\bibitem[Shirzad et~al.(2023)Shirzad, Velingker, Venkatachalam, Sutherland, and Sinop]{shirzad2023exphormer}
Shirzad, H., Velingker, A., Venkatachalam, B., Sutherland, D.~J., and Sinop, A.~K.
\newblock Exphormer: Sparse transformers for graphs.
\newblock In \emph{International Conference on Machine Learning}, pp.\  31613--31632. PMLR, 2023.

\bibitem[Shoeybi et~al.(2019)Shoeybi, Patwary, Puri, LeGresley, Casper, and Catanzaro]{shoeybi2019megatron}
Shoeybi, M., Patwary, M., Puri, R., LeGresley, P., Casper, J., and Catanzaro, B.
\newblock Megatron-lm: Training multi-billion parameter language models using model parallelism.
\newblock \emph{CoRR}, abs/1909.08053, 2019.
\newblock URL \url{http://arxiv.org/abs/1909.08053}.

\bibitem[Sun et~al.(2025)Sun, Zhang, Hu, Gu, Chen, and Yang]{sun2020adaptive}
Sun, C., Zhang, M., Hu, J., Gu, H., Chen, J., and Yang, M.
\newblock Adaptive graph diffusion networks: compact and expressive gnns with large receptive fields.
\newblock \emph{Artif. Intell. Rev.}, 58\penalty0 (4):\penalty0 107, 2025.
\newblock \doi{10.1007/S10462-025-11114-Z}.
\newblock URL \url{https://doi.org/10.1007/s10462-025-11114-z}.

\bibitem[Sun et~al.(2023)Sun, Dou, Yang, Zhang, Wang, Yu, He, and Li]{sun2023adversarial}
Sun, L., Dou, Y., Yang, C., Zhang, K., Wang, J., Yu, P.~S., He, L., and Li, B.
\newblock Adversarial attack and defense on graph data: {A} survey.
\newblock \emph{{IEEE} Trans. Knowl. Data Eng.}, 35\penalty0 (8):\penalty0 7693--7711, 2023.
\newblock \doi{10.1109/TKDE.2022.3201243}.
\newblock URL \url{https://doi.org/10.1109/TKDE.2022.3201243}.

\bibitem[Szklarczyk et~al.(2019)Szklarczyk, Gable, Lyon, Junge, Wyder, Huerta-Cepas, Simonovic, Doncheva, Morris, Bork, et~al.]{szklarczyk2019string}
Szklarczyk, D., Gable, A.~L., Lyon, D., Junge, A., Wyder, S., Huerta-Cepas, J., Simonovic, M., Doncheva, N.~T., Morris, J.~H., Bork, P., et~al.
\newblock String v11: protein--protein association networks with increased coverage, supporting functional discovery in genome-wide experimental datasets.
\newblock \emph{Nucleic acids research}, 47\penalty0 (D1):\penalty0 D607--D613, 2019.

\bibitem[Touvron et~al.(2023)Touvron, Lavril, Izacard, Martinet, Lachaux, Lacroix, Rozi{\`e}re, Goyal, Hambro, Azhar, et~al.]{touvron2023llama}
Touvron, H., Lavril, T., Izacard, G., Martinet, X., Lachaux, M.-A., Lacroix, T., Rozi{\`e}re, B., Goyal, N., Hambro, E., Azhar, F., et~al.
\newblock Llama: Open and efficient foundation language models.
\newblock \emph{arXiv preprint arXiv:2302.13971}, 2023.

\bibitem[Vaswani et~al.(2017)Vaswani, Shazeer, Parmar, Uszkoreit, Jones, Gomez, Kaiser, and Polosukhin]{vaswani2017attention}
Vaswani, A., Shazeer, N., Parmar, N., Uszkoreit, J., Jones, L., Gomez, A.~N., Kaiser, {\L}., and Polosukhin, I.
\newblock Attention is all you need.
\newblock In \emph{Advances in Neural Information Processing Systems}, pp.\  5998--6008, 2017.

\bibitem[Wang et~al.(2019)Wang, Singh, Michael, Hill, Levy, and Bowman]{wang2019glue}
Wang, A., Singh, A., Michael, J., Hill, F., Levy, O., and Bowman, S.~R.
\newblock {GLUE:} {A} multi-task benchmark and analysis platform for natural language understanding.
\newblock In \emph{7th International Conference on Learning Representations, {ICLR} 2019, New Orleans, LA, USA, May 6-9, 2019}. OpenReview.net, 2019.
\newblock URL \url{https://openreview.net/forum?id=rJ4km2R5t7}.

\bibitem[Wang et~al.(2023)Wang, Fu, Du, Gao, Huang, Liu, Chandak, Liu, Katwyk, Deac, Anandkumar, Bergen, Gomes, Ho, Kohli, Lasenby, Leskovec, Liu, Manrai, Marks, Ramsundar, Song, Sun, Tang, Velickovic, Welling, Zhang, Coley, Bengio, and Zitnik]{wang2023ai4sci}
Wang, H., Fu, T., Du, Y., Gao, W., Huang, K., Liu, Z., Chandak, P., Liu, S., Katwyk, P.~V., Deac, A., Anandkumar, A., Bergen, K., Gomes, C.~P., Ho, S., Kohli, P., Lasenby, J., Leskovec, J., Liu, T., Manrai, A., Marks, D.~S., Ramsundar, B., Song, L., Sun, J., Tang, J., Velickovic, P., Welling, M., Zhang, L., Coley, C.~W., Bengio, Y., and Zitnik, M.
\newblock Scientific discovery in the age of artificial intelligence.
\newblock \emph{Nat.}, 620\penalty0 (7972):\penalty0 47--60, 2023.
\newblock \doi{10.1038/S41586-023-06221-2}.
\newblock URL \url{https://doi.org/10.1038/s41586-023-06221-2}.

\bibitem[Wei et~al.(2022)Wei, Bosma, Zhao, Guu, Yu, Lester, Du, Dai, and Le]{wei2021finetuned}
Wei, J., Bosma, M., Zhao, V.~Y., Guu, K., Yu, A.~W., Lester, B., Du, N., Dai, A.~M., and Le, Q.~V.
\newblock Finetuned language models are zero-shot learners.
\newblock In \emph{The Tenth International Conference on Learning Representations, {ICLR} 2022, Virtual Event, April 25-29, 2022}. OpenReview.net, 2022.
\newblock URL \url{https://openreview.net/forum?id=gEZrGCozdqR}.

\bibitem[West et~al.(2001)]{west2001introduction}
West, D.~B. et~al.
\newblock \emph{Introduction to graph theory}, volume~2.
\newblock Prentice hall Upper Saddle River, 2001.

\bibitem[Wolf et~al.(2020)Wolf, Debut, Sanh, Chaumond, Delangue, Moi, Cistac, Rault, Louf, Funtowicz, Davison, Shleifer, von Platen, Ma, Jernite, Plu, Xu, Scao, Gugger, Drame, Lhoest, and Rush]{wolf-etal-2020-transformers}
Wolf, T., Debut, L., Sanh, V., Chaumond, J., Delangue, C., Moi, A., Cistac, P., Rault, T., Louf, R., Funtowicz, M., Davison, J., Shleifer, S., von Platen, P., Ma, C., Jernite, Y., Plu, J., Xu, C., Scao, T.~L., Gugger, S., Drame, M., Lhoest, Q., and Rush, A.~M.
\newblock Transformers: State-of-the-art natural language processing.
\newblock In \emph{Proceedings of the 2020 Conference on Empirical Methods in Natural Language Processing: System Demonstrations}, pp.\  38--45, Online, October 2020. Association for Computational Linguistics.
\newblock URL \url{https://www.aclweb.org/anthology/2020.emnlp-demos.6}.

\bibitem[Wu et~al.(2022)Wu, Zhao, Li, Wipf, and Yan]{wu2022nodeformer}
Wu, Q., Zhao, W., Li, Z., Wipf, D.~P., and Yan, J.
\newblock Nodeformer: A scalable graph structure learning transformer for node classification.
\newblock \emph{Advances in Neural Information Processing Systems}, 35:\penalty0 27387--27401, 2022.

\bibitem[Wu et~al.(2024)Wu, Zhao, Yang, Zhang, Nie, Jiang, Bian, and Yan]{wu2024sgformer}
Wu, Q., Zhao, W., Yang, C., Zhang, H., Nie, F., Jiang, H., Bian, Y., and Yan, J.
\newblock Simplifying and empowering transformers for large-graph representations.
\newblock \emph{Advances in Neural Information Processing Systems}, 36, 2024.

\bibitem[Wu et~al.(2017)Wu, Ramsundar, Feinberg, Gomes, Geniesse, Pappu, Leswing, and Pande]{wu2017molecule}
Wu, Z., Ramsundar, B., Feinberg, E.~N., Gomes, J., Geniesse, C., Pappu, A.~S., Leswing, K., and Pande, V.~S.
\newblock Moleculenet: {A} benchmark for molecular machine learning.
\newblock \emph{CoRR}, abs/1703.00564, 2017.
\newblock URL \url{http://arxiv.org/abs/1703.00564}.

\bibitem[Wu et~al.(2020)Wu, Pan, Chen, Long, Zhang, and Philip]{wu2020comprehensive}
Wu, Z., Pan, S., Chen, F., Long, G., Zhang, C., and Philip, S.~Y.
\newblock A comprehensive survey on graph neural networks.
\newblock \emph{IEEE transactions on neural networks and learning systems}, 32\penalty0 (1):\penalty0 4--24, 2020.

\bibitem[Xu et~al.(2019)Xu, Hu, Leskovec, and Jegelka]{xu2019how}
Xu, K., Hu, W., Leskovec, J., and Jegelka, S.
\newblock How powerful are graph neural networks?
\newblock In \emph{7th International Conference on Learning Representations, {ICLR} 2019, New Orleans, LA, USA, May 6-9, 2019}. OpenReview.net, 2019.
\newblock URL \url{https://openreview.net/forum?id=ryGs6iA5Km}.

\bibitem[Yang et~al.(2023)Yang, Feng, Shen, and Hooi]{yang2023pdf}
Yang, M., Feng, W., Shen, Y., and Hooi, B.
\newblock Towards better graph representation learning with parameterized decomposition {\&} filtering.
\newblock In Krause, A., Brunskill, E., Cho, K., Engelhardt, B., Sabato, S., and Scarlett, J. (eds.), \emph{International Conference on Machine Learning, {ICML} 2023, 23-29 July 2023, Honolulu, Hawaii, {USA}}, volume 202 of \emph{Proceedings of Machine Learning Research}, pp.\  39234--39251. {PMLR}, 2023.
\newblock URL \url{https://proceedings.mlr.press/v202/yang23c.html}.

\bibitem[Ying et~al.(2021)Ying, Cai, Luo, Zheng, Ke, He, Shen, and Liu]{ying2021transformers}
Ying, C., Cai, T., Luo, S., Zheng, S., Ke, G., He, D., Shen, Y., and Liu, T.-Y.
\newblock Do transformers really perform badly for graph representation?
\newblock \emph{Advances in Neural Information Processing Systems}, 34:\penalty0 28877--28888, 2021.

\bibitem[Zeng et~al.(2021)Zeng, Zhang, Xia, Srivastava, Malevich, Kannan, Prasanna, Jin, and Chen]{zeng2021decoupling}
Zeng, H., Zhang, M., Xia, Y., Srivastava, A., Malevich, A., Kannan, R., Prasanna, V., Jin, L., and Chen, R.
\newblock Decoupling the depth and scope of graph neural networks.
\newblock \emph{Advances in Neural Information Processing Systems}, 34:\penalty0 19665--19679, 2021.

\bibitem[Zhang et~al.(2020)Zhang, Zhang, Xia, and Sun]{zhang2020graphbert}
Zhang, J., Zhang, H., Xia, C., and Sun, L.
\newblock Graph-bert: Only attention is needed for learning graph representations.
\newblock \emph{CoRR}, abs/2001.05140, 2020.
\newblock URL \url{https://arxiv.org/abs/2001.05140}.

\bibitem[Zhang et~al.(2023)Zhang, Yan, He, Li, and Chu]{zhang2023drgcn}
Zhang, L., Yan, X., He, J., Li, R., and Chu, W.
\newblock {DRGCN:} dynamic evolving initial residual for deep graph convolutional networks.
\newblock In Williams, B., Chen, Y., and Neville, J. (eds.), \emph{Thirty-Seventh {AAAI} Conference on Artificial Intelligence, {AAAI} 2023, Thirty-Fifth Conference on Innovative Applications of Artificial Intelligence, {IAAI} 2023, Thirteenth Symposium on Educational Advances in Artificial Intelligence, {EAAI} 2023, Washington, DC, USA, February 7-14, 2023}, pp.\  11254--11261. {AAAI} Press, 2023.
\newblock \doi{10.1609/AAAI.V37I9.26332}.
\newblock URL \url{https://doi.org/10.1609/aaai.v37i9.26332}.

\bibitem[Zhang \& Chen(2018)Zhang and Chen]{zhang2018link}
Zhang, M. and Chen, Y.
\newblock Link prediction based on graph neural networks.
\newblock In Bengio, S., Wallach, H.~M., Larochelle, H., Grauman, K., Cesa{-}Bianchi, N., and Garnett, R. (eds.), \emph{Advances in Neural Information Processing Systems 31: Annual Conference on Neural Information Processing Systems 2018, NeurIPS 2018, December 3-8, 2018, Montr{\'{e}}al, Canada}, pp.\  5171--5181, 2018.

\bibitem[Zhang \& Li(2021)Zhang and Li]{zhang2021nestedgnn}
Zhang, M. and Li, P.
\newblock Nested graph neural networks.
\newblock In Ranzato, M., Beygelzimer, A., Dauphin, Y.~N., Liang, P., and Vaughan, J.~W. (eds.), \emph{Advances in Neural Information Processing Systems 34: Annual Conference on Neural Information Processing Systems 2021, NeurIPS 2021, December 6-14, 2021, virtual}, pp.\  15734--15747, 2021.

\bibitem[Zhang et~al.(2021)Zhang, Li, Xia, Wang, and Jin]{zhang2021labeling}
Zhang, M., Li, P., Xia, Y., Wang, K., and Jin, L.
\newblock Labeling trick: {A} theory of using graph neural networks for multi-node representation learning.
\newblock In Ranzato, M., Beygelzimer, A., Dauphin, Y.~N., Liang, P., and Vaughan, J.~W. (eds.), \emph{Advances in Neural Information Processing Systems 34: Annual Conference on Neural Information Processing Systems 2021, NeurIPS 2021, December 6-14, 2021, virtual}, pp.\  9061--9073, 2021.

\bibitem[Zhou et~al.(2009)Zhou, L{\"u}, and Zhang]{zhou2009predicting}
Zhou, T., L{\"u}, L., and Zhang, Y.-C.
\newblock Predicting missing links via local information.
\newblock \emph{The European Physical Journal B}, 71:\penalty0 623--630, 2009.

\end{thebibliography}
\bibliographystyle{icml2025}

\newpage
\appendix
\onecolumn

\section{Datasets}
\label{app:data}
The detailed statistics of the datasets are in the Tab. \ref{tab:stats}.

\begin{table}[ht]
\caption{Statistics of graph-/edge-/node-level datasets. Here `BC' stands for binary classification. $p$ in Random Graph datasets means the probability of 
creating the edge between a node pair.}
\label{tab:stats}
\vskip 0.15in
\begin{center}
\begin{small}
\begin{tabular}{l|l|l|l|l|l}
\toprule
datasets      & \# of graphs & avg \# of nodes & avg \# of edges & task-type                        & metrics \\
\midrule
PCQM4Mv2         & 3,746,619    & 14.14            & 14.56       & regression     & MAE     \\
ogbg-molpcba     & 437,929      & 26.0             & 28.1        & multi-label BC & AP \\
reddit-threads   & 203,088      & 23.9             & 24.9        & BC             & ROC-AUC \\
Triangles        & 45,000       & 20.9             & 32.7        & multi-class classification & ACC \\
Internal dataset & 3,100,000    & 24.8             & 54.7       & N/A & N/A \\
Random Graph$_{p=0.03}$     & 3,100,000    & 67.1           & 74.8        & N/A & N/A \\
\midrule
ogbl-ppa         & 1            & 576,289          & 30,326,273  &         BC     & HR@100  \\
ogbl-citation2   & 1            & 2,927,963        & 30,561,187  &         BC     & MRR  \\
\midrule
ogbn-proteins    & 1            & 132,534          & 39,561,252  & multi-label BC &  ROC-AUC \\ 
ogbn-arxiv       & 1            & 169,343          & 1,166,243   & multi-class classification &  ACC \\ 
\bottomrule
\end{tabular}
\end{small}
\end{center}
\vskip -0.1in
\end{table}

\subsection{Subgraph Sampling}
\label{app:subgraph}
The subgraph sampling configurations for different datasets of large graphs are shown in the Tab. \ref{tab:sampling}.

\begin{table}[ht]
\caption{Details of subgraph sampling for ogbl-ppa and ogbn-proteins datasets. `seq-len' means the average length of the Eulerian sequences. Edge-ego means sampling subgraph around the central edge, and Node-ego means sampling around the central node.}
\label{tab:sampling}
\vskip 0.15in
\begin{center}
\begin{small}
\begin{tabular}{l|l|l|l|l}
\toprule
dataset                        & sampling                  & depth ($d$) & \# neighbors ($n$) & seq-len  \\
\midrule
\multirow{2}{*}{ogbl-ppa}      & \multirow{2}{*}{edge-ego} & 1     & 14           & 90  \\
                               &                           & 1     & 30           & 280  \\
\midrule
\multirow{3}{*}{ogbl-citation2} & \multirow{3}{*}{edge-ego} & 1    & 14           & 60  \\
                               &                           & 1     & 20           & 90  \\
                               &                           & 1     & 30           & 130  \\
\midrule
\multirow{4}{*}{ogbn-proteins} & \multirow{4}{*}{node-ego} & 9     & 1            & 20  \\
                               &                           & 20    & 1            & 50  \\
                               &                           & 40    & 1            & 120 \\
                               &                           & 60    & 1            & 200 \\
\midrule
\multirow{2}{*}{ogbn-arxiv} & \multirow{2}{*}{node-ego}     & 1     & 30            & 30  \\
                               &                            & 1    & 40            & 40 \\
\bottomrule
\end{tabular}
\end{small}
\end{center}
\vskip -0.1in
\end{table}

\section{Models}
We list the model specifics in the Tab. \ref{tab:models}. We experiment with eight different scales of models.

\begin{table}[ht]
\caption{Statistics of GraphGPT models of different sizes. The GraphGPT-Base is of the same scale as Bert-Base \citep{devlin2018bert}.
}
\label{tab:models}
\vskip 0.15in
\begin{center}
\begin{small}
\begin{tabular}{l|l|l|l|l}
\toprule
Model-size    & Hidden-size & \# of layers & \# of heads  & Params (excluding embed)\\ 
\midrule
Mini                 & 256         & 4            & 4     & 4.2M          \\
S (Small)            & 512         & 4            & 8     & 16.8M         \\
M (Medium)           & 512         & 8            & 8     & 33.6M         \\
B / B$_{12}$ (Base)  & 768         & 12           & 12    & 113.2M          \\
B$_{24}$ (Base24)    & 768         & 24           & 12    & 226.5M          \\
B$_{48}$ (Base48)    & 768         & 48           & 12    & 453.0M           \\
L (Large)            & 1024        & 24           & 16    & 402.7M        \\
XXL (XXLarge)        & 1600        & 48           & 25    & 2.0B              \\ 
\bottomrule
\end{tabular}
\end{small}
\end{center}
\vskip -0.1in
\end{table}

\section{Implementation Details}
\label{app:impl}
\subsection{Graphs to Sequences of Tokens}
The implementation uses PyTorch as the primary framework. For graph preprocessing tasks such as subgraph sampling, we utilize torch-geometric \citep{fey2019}. When required, we employ NetworkX \citep{hagberg2008exploring} to Eulerize (sub)graphs and identify (semi-)Eulerian paths. A custom tokenizer converts these paths into token sequences, with dataset-specific vocabularies constructed for each case.

\subsection{Model Backbone}
We employ a transformer architecture based on Llama \citep{touvron2023llama}, implemented via the Hugging Face Transformers library \citep{wolf-etal-2020-transformers}, as the backbone for NTP pre-training. For SMTP pre-training, we modify the architecture by replacing the causal attention mask with a bidirectional attention mask to create an encoder. We initialize all parameters randomly, and train models at various scales (see Table \ref{tab:models}).

\subsection{Training}
The models are pre-trained and fine-tuned on A800-80G GPU clusters\footnote{We also utilize clusters of other types of GPUs, for example, Nvidia V100-32G, L20, L40 and etc.} using DeepSpeed's Stage-2 strategy with mixed precision (FP16/FP32) or BF16 \citep{rasley2020deepspeed}. We employ the AdamW optimizer \citep{loshchilov2017decoupled} with a learning rate scheduler. To maximize computational efficiency in pre-training stage, we pack multiple graph sequences into single entries, optimizing context window utilization \citep{raffel2020exploring} for certain graph datasets. Dataset-specific configurations are detailed in their respective sections below.

The variance is inherently low for most large-scale datasets (e.g., PCQM4Mv2, ogbl-ppa), where reporting variance is standard practice only when significant. For these datasets, 3–5 runs consistently yield minimal variance (as shown in tables). For the Triangles dataset, variance is higher—particularly on out-of-distribution (OOD) test data. So we conducted 10 runs to ensure robustness.

\begin{figure}[htpb]
\begin{center}
\includegraphics[width=\textwidth]{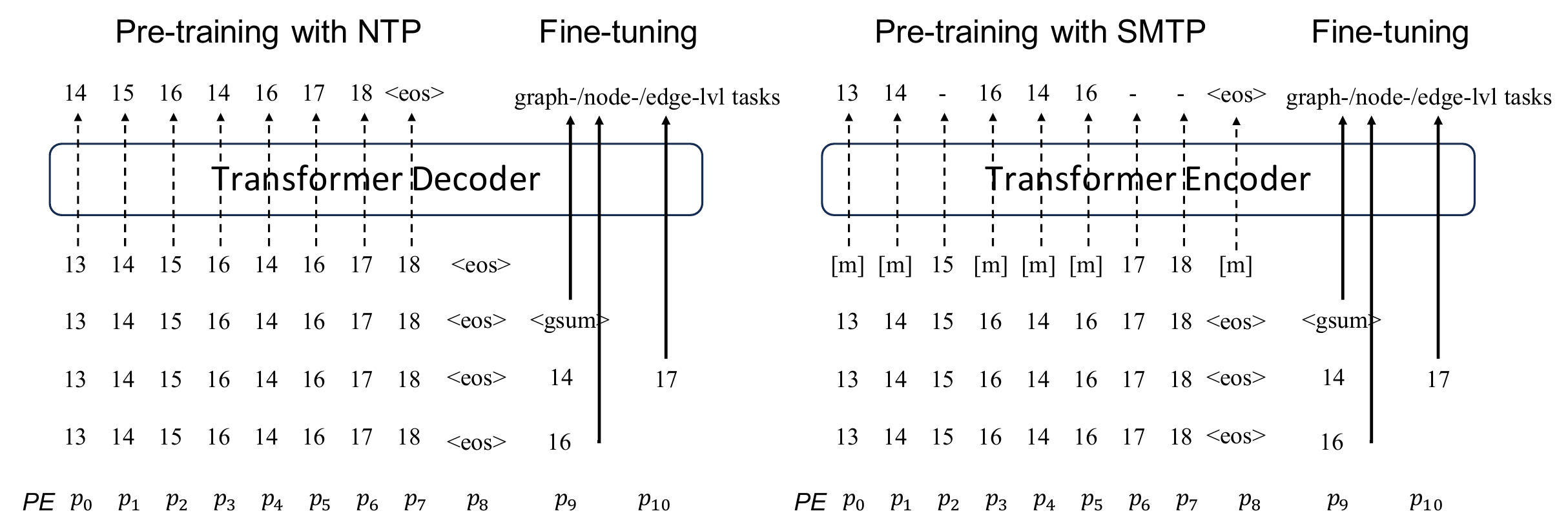}
\end{center}
\caption{Pre-training and fine-tuning illustrations.}
\label{fig:pre-train-fine-tune}
\end{figure}

The pre-training and fine-tuning paradigm is illustrated in Fig. \ref{fig:pre-train-fine-tune}.

\subsection{Vocabulary}
\label{app:vocabulary}
In NLP, the vocabulary is typically constructed by tokenizing text data using the byte-pair encoding (BPE) algorithm \citep{sennrich2016neural}. The resulting unique tokens form the vocabulary, which usually comprises frequent subwords from the text corpus.

In contrast, our GraphGPT employs a fundamentally different vocabulary construction approach. The vocabulary is split into two distinct parts: \textit{1). structural and special tokens}, which are dataset-agnostic and transferable across different datasets. \textit{2). semantic tokens}, which encode dataset-specific information, such as node and edge attributes.

An example is provided in Appendix \ref{app:edge}. In the graph sequence: \textit{i).} tokens like `1', `2', etc., represent \textit{structural tokens}; \textit{ii).} Tokens such as `ogbl-ppa\#node\#0\#17' and `ogbl-ppa\#node\#1\#1959' are \textit{semantics tokens}; \textit{iii).} \textit{Special tokens} like $<$gsum$>$ and $<$eos$>$ (Fig. \ref{fig:eulerize}) denote graph-specific functions (e.g., graph summary and end-of-sequence markers).

For the datasets ogbl-ppa/citation2, ogbn-proteins/arxiv, we set $k=2$ (\S \ref{sec:node-coding}), resulting in vocabulary sizes of $41,634$ / $25,687$ and $31,360$ / $25,600$, respectively.

\section{Graphs to Sequences of Tokens}

In this section, we show some examples of turning graphs to sequences of tokens.

\subsection{Molecular Graphs to Tokens}

Below is one example of 2D molecular graphs in the ogbg-molpcba dataset in torch-geometric data format \citep{fey2019}.

\begin{lstlisting}[breaklines]
Data(x=[4, 9], edge_index=[2, 6], edge_attr=[6, 3], y=[128])
\end{lstlisting}

The graph has 4 nodes and 3 edges. The source and destination nodes of the edges are  recorded in `edge\_index', and its dimension is ($2, 2\cdot\text{number\_of\_edges}$) for undirected graphs. `x' is the node attributes of 9 dimensions, and `edge\_attr' stores the edge attributes of 3 dimensions. 

The node and edge attributes of the graphs are numbers. If we directly discretize them into tokens, i.e., using one token to represent each unique number, the numbers that appear few times in the dataset cannot be well-trained. At the same time, the vocabulary may blow up. Therefore, we split them into single digits and represent them with the combination of the following tokens. They are dataset agnostic, and can be shared across different datasets.
\begin{lstlisting}[breaklines]
<->, <.>, <0>, <1>, <2>, <3>, <4>, <5>, <6>, <7>, <8>, <9>
\end{lstlisting}

The resulting vocabulary is 556 for both ogbg-molpcba and PCQM4Mv2.

Below shows the tokens from one of the possible (semi-)Eulerain paths of the above molecular graph.
\begin{lstlisting}[breaklines]
['1', 'ogbg-molpcba#node#0#1', '<7>', 'ogbg-molpcba#node#2#1', '<1>', 'ogbg-molpcba#node#3#1', '<5>', 'ogbg-molpcba#node#6#1', '<1>', 'ogbg-molpcba#edge#0#1', '<1>', '2', '3', 'ogbg-molpcba#node#0#1', '<5>', 'ogbg-molpcba#node#2#1', '<4>', 'ogbg-molpcba#node#3#1', '<5>', 'ogbg-molpcba#node#4#1', '<3>', 'ogbg-molpcba#node#6#1', '<2>', '2', 'ogbg-molpcba#node#0#1', '<5>', 'ogbg-molpcba#node#2#1', '<3>', 'ogbg-molpcba#node#3#1', '<5>', 'ogbg-molpcba#node#6#1', '<1>', '4', 'ogbg-molpcba#node#0#1', '<5>', 'ogbg-molpcba#node#2#1', '<4>', 'ogbg-molpcba#node#3#1', '<5>', 'ogbg-molpcba#node#4#1', '<3>', 'ogbg-molpcba#node#6#1', '<2>']
\end{lstlisting}

In the sequence of tokens above, for the node `1', we can deduce that its 9 dimensional attributes are $(7, 0, 1, 5, 0, 0, 1, 0, 0, 0)$. Node `1' is connected to `2' with edge attributes $(1, 0, 0)$. We set 0 as the the default value of the attributes in this dataset, and do not encode it into tokens.

In the (semi-)Eulerian path, a node may appear several times. We append its attributes tokens to one of its appearances randomly. This can prevent the model from copying the attributes from the previous appearance, and also shorten the resulting sequence.

For a graph obtained from Eulerization, an edge may present several times in the path. We apply the same logic to insert the edge  attributes tokens.

As in the above sequence, node `2' appears two times, and its node attributes tokens are appended after its second appearance. There is no tokens encode the edge attributes of edge between `2' and `3', which implies the edge attributes are default value $(0, 0, 0)$.


\subsection{Subgraphs to Tokens}
In edge/node-level tasks, we usually have one big graph. In this section, we use ogbl-ppa and ogbn-proteins datasets to 
show how to sample subgraphs from the big graph, and then transform the subgraph to sequences of tokens.

The whole ogbl-ppa dataset is summarized in torch-geometric format as follows.

\begin{lstlisting}[breaklines]
Data(num_nodes=576289, edge_index=[2, 42463862], x=[576289, 58])
\end{lstlisting}

It has 576289 nodes and 21231931 edges in the training data. `x' is the one-hot representation of the species that the node (protein) belongs to.

We sample a subgraph from it as below.
\begin{lstlisting}[breaklines]
Data(num_nodes=30, root_n_id=[2], edge_index=[2, 84], x=[30, 2])
\end{lstlisting}

It has 30 nodes, 42 edges as in `edge\_index'. `x' is the node attributes of 2 dimensions, and it encodes the node identity as described in Sec. \ref{sec:node-coding}. We partition the nodes (proteins) based on the associated species. The number of proteins inside each species varies from 616 to 41017. Finally we use 58 tokens for species and 41017 tokens for the local indices. Combined with the tokens for the structure and the special tokens, the total vocabulary is 41231.

Here `root\_n\_id' records the two seed nodes, and the subgraph is sampled centered around them. The resulting tokens from one of the possible (semi-)Eulerian paths are:
\begin{lstlisting}[breaklines]
['1', '2', '3', 'ogbl-ppa#node#0#17', 'ogbl-ppa#node#1#1959', '4', '5', 'ogbl-ppa#node#0#17', 'ogbl-ppa#node#1#2460', '6', '7', 'ogbl-ppa#node#0#17', 'ogbl-ppa#node#1#3566', '6', '8', 'ogbl-ppa#node#0#17', 'ogbl-ppa#node#1#4145', '6', '9', 'ogbl-ppa#node#0#20', 'ogbl-ppa#node#1#5334', '10', 'ogbl-ppa#node#0#27', 'ogbl-ppa#node#1#17324', '6', 'ogbl-ppa#node#0#17', 'ogbl-ppa#node#1#6850', '11', 'ogbl-ppa#node#0#17', 'ogbl-ppa#node#1#5498', '6', '12', 'ogbl-ppa#node#0#17', 'ogbl-ppa#node#1#5776', '6', '4', 'ogbl-ppa#node#0#17', 'ogbl-ppa#node#1#8183', '2', '5', '2', '13', 'ogbl-ppa#node#0#17', 'ogbl-ppa#node#1#3514', '2', 'ogbl-ppa#node#0#17', 'ogbl-ppa#node#1#9374', '14', 'ogbl-ppa#node#0#17', 'ogbl-ppa#node#1#6164', '15', 'ogbl-ppa#node#0#17', 'ogbl-ppa#node#1#8368', '2', '6', '16', 'ogbl-ppa#node#0#17', 'ogbl-ppa#node#1#10803', '6', '17', 'ogbl-ppa#node#0#17', 'ogbl-ppa#node#1#11465', '6', '10', '18', 'ogbl-ppa#node#0#20', 'ogbl-ppa#node#1#16505', '6', '19', 'ogbl-ppa#node#0#17', 'ogbl-ppa#node#1#15071', '2', '20', 'ogbl-ppa#node#0#17', 'ogbl-ppa#node#1#7761', '2', '21', 'ogbl-ppa#node#0#17', 'ogbl-ppa#node#1#8828', '2', '22', 'ogbl-ppa#node#0#17', 'ogbl-ppa#node#1#14477', '2', '23', 'ogbl-ppa#node#0#17', 'ogbl-ppa#node#1#16026', '2', '24', 'ogbl-ppa#node#0#17', 'ogbl-ppa#node#1#16825', '6', '25', 'ogbl-ppa#node#0#17', 'ogbl-ppa#node#1#17615', '19', '25', '2', '26', 'ogbl-ppa#node#0#17', 'ogbl-ppa#node#1#19524', '2', '27', 'ogbl-ppa#node#0#17', 'ogbl-ppa#node#1#17854', '6', '28', 'ogbl-ppa#node#0#17', 'ogbl-ppa#node#1#17733', '6', '29', 'ogbl-ppa#node#0#27', 'ogbl-ppa#node#1#23255', '6', '30', 'ogbl-ppa#node#0#17', 'ogbl-ppa#node#1#19700', '6', '27', '1', 'ogbl-ppa#node#0#17', 'ogbl-ppa#node#1#20474']
\end{lstlisting}

In the ablation study on node identity encoding in Sec. \ref{sec:ablation-node-id}, an example of the subgraph sampled from ogbl-ppa without identity encoding is shown below.
\begin{lstlisting}[breaklines]
Data(num_nodes=30, root_n_id=[2], edge_index=[2, 136], x=[30, 1])
\end{lstlisting}

Different from the subgraph with node identity encoded in `x', its node attribute `x' contains only the information of the node's (protein) hosting species. It cannot be used to uniquely identify the nodes. The vocabulary decreases from 41231 to 214.

The resulting tokens from one of its possible (semi-)Eulerian paths is below.
\begin{lstlisting}[breaklines]
['1', '2', '3', '4', 'ogbl-ppa#node#0#17', '5', '6', '7', '5', '8', '9', '1', 'ogbl-ppa#node#0#17', '10', 'ogbl-ppa#node#0#17', '11', 'ogbl-ppa#node#0#17', '3', 'ogbl-ppa#node#0#17', '11', '12', '1', '5', '13', 'ogbl-ppa#node#0#17', '5', '14', 'ogbl-ppa#node#0#17', '5', '9', '10', '8', 'ogbl-ppa#node#0#17', '3', '15', 'ogbl-ppa#node#0#17', '3', '16', 'ogbl-ppa#node#0#17', '3', '2', 'ogbl-ppa#node#0#20', '17', 'ogbl-ppa#node#0#27', '1', '18', 'ogbl-ppa#node#0#20', '1', '19', 'ogbl-ppa#node#0#17', '3', '9', 'ogbl-ppa#node#0#17', '20', 'ogbl-ppa#node#0#17', '10', '3', '21', '3', '5', '10', '12', 'ogbl-ppa#node#0#17', '3', '22', 'ogbl-ppa#node#0#17', '3', '17', '18', '3', '23', '13', '24', '5', '25', 'ogbl-ppa#node#0#17', '23', 'ogbl-ppa#node#0#17', '21', 'ogbl-ppa#node#0#17', '20', '5', '26', 'ogbl-ppa#node#0#17', '5', '22', '24', 'ogbl-ppa#node#0#17', '23', '5', '27', '6', 'ogbl-ppa#node#0#17', '28', 'ogbl-ppa#node#0#17', '7', 'ogbl-ppa#node#0#17', '28', '5', 'ogbl-ppa#node#0#17', '27', 'ogbl-ppa#node#0#17', '29', 'ogbl-ppa#node#0#17', '5', '30', 'ogbl-ppa#node#0#17', '5', '19', '5', '12', '20', '1']
\end{lstlisting}

In the following, we use the ogbn-proteins dataset as the example. The entire dataset is a large graph as below. 

\begin{lstlisting}[breaklines]
Data(num_nodes=132534, edge_index=[2, 79122504], edge_attr=[79122504, 8], node_species=[132534, 1], y=[132534, 112])
\end{lstlisting}

It has 132,534 nodes and 39,561,252 edges. `node\_species' stores the species' numeric id that the node (proteins) belongs to.

One sampled subgraph in the torch-geometric data format is:
\begin{lstlisting}[breaklines]
Data(num_nodes=10, root_n_id=0, edge_index=[2, 22], edge_attr=[22, 8], y=[10, 112], x=[10, 2])
\end{lstlisting}

It has 10 nodes, 11 edges as in `edge\_index'. Edge attributes is stored in `edge\_attr' of dimension 8. `x' is the node attributes of 2 dimensions, and it encodes the node identity as described in Sec. \ref{sec:node-coding}. Its first dimension (token) represents the species, and the second is local numbering of each protein inside its species. Similar to the ogbl-ppa dataset, the identity encoding of 132,534 nodes occupies 25,465 tokens in the vocabulary, and the total vocabulary is 25,620.

`y' records the labels for the supervised node-level task. `root\_n\_id' represents the target node, and the subgraph is sampled centered around it.

The resulting tokens from one of the possible (semi-)Eulerian paths are as follows.
\begin{lstlisting}[breaklines]
['1', 'ogbn-proteins#node#0#3702', 'ogbn-proteins#node#1#16267', 'ogbn-proteins#edge#7#1', '<1>', '<6>', '<4>', '2', 'ogbn-proteins#node#0#3702', 'ogbn-proteins#node#1#6896', 'ogbn-proteins#edge#4#1', '<3>', '<4>', '<0>', '3', 'ogbn-proteins#node#0#3702', 'ogbn-proteins#node#1#4121', 'ogbn-proteins#edge#4#1', '<3>', '<9>', '<8>', '4', 'ogbn-proteins#node#0#3702', 'ogbn-proteins#node#1#3963', 'ogbn-proteins#edge#4#1', '<1>', '<5>', '<3>', '5', 'ogbn-proteins#node#0#3702', 'ogbn-proteins#node#1#8259', 'ogbn-proteins#edge#4#1', '<4>', '<8>', 'ogbn-proteins#edge#7#1', '<2>', '<1>', '<5>', '6', '7', 'ogbn-proteins#edge#7#1', '<4>', '<1>', '<8>', '8', 'ogbn-proteins#node#0#3702', 'ogbn-proteins#node#1#1', '7', 'ogbn-proteins#node#0#3702', 'ogbn-proteins#node#1#89', 'ogbn-proteins#edge#7#1', '<3>', '<2>', '<1>', '6', 'ogbn-proteins#node#0#3702', 'ogbn-proteins#node#1#955', 'ogbn-proteins#edge#7#1', '<2>', '<7>', '<0>', '9', 'ogbn-proteins#node#0#3702', 'ogbn-proteins#node#1#7055', 'ogbn-proteins#edge#4#1', '<1>', '<6>', '<5>', '10', 'ogbn-proteins#node#0#3702', 'ogbn-proteins#node#1#10010', 'ogbn-proteins#edge#4#1', '<1>', '<6>', '<9>', '4', '5', 'ogbn-proteins#edge#4#1', '<2>', '<0>', '<7>', '3']
\end{lstlisting}

The original edge attributes are 8-dimensional vector of 3 decimal numbers from $0.001$ to $1$. We split them into single digits and represent them with the combination of the digits tokens as in App. \ref{app:graph}.

To reduce the number of tokens in the resultant sequences further, we multiply the number with 1000 and then minus it by 1. So we do not need to encode `.' any more. At the same time, we treat the value $0.001$ ($0$ after the above transformation) as the default value and do not encode it with tokens.

\section{Graph-Level Task}
\label{app:graph}

\subsection{PCQM4M-v2}


\begin{figure}[htpb]
\begin{center}
    \begin{subfigure}{0.45\textwidth}
        \includegraphics[width=\textwidth]{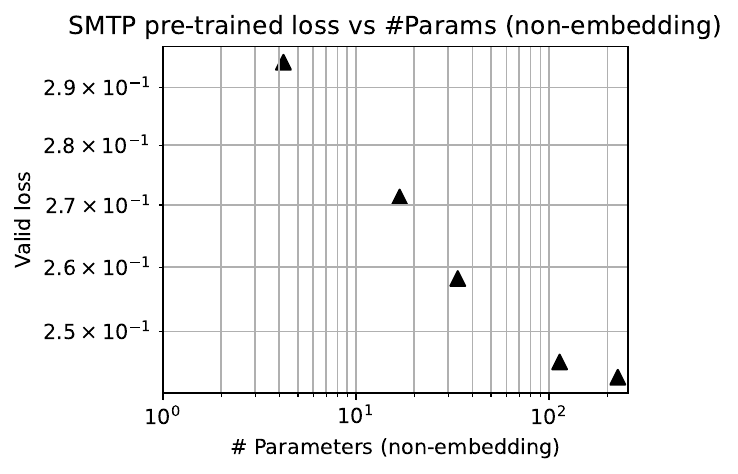}
    \end{subfigure}
    \hfill
    \begin{subfigure}{0.45\textwidth}
        \includegraphics[width=\textwidth]{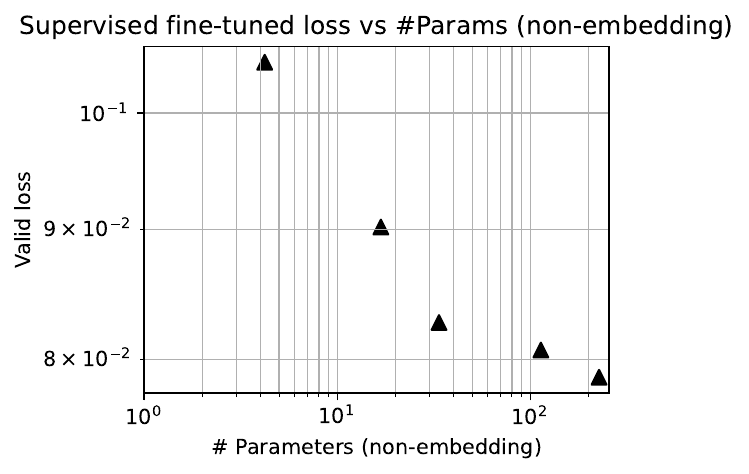}
    \end{subfigure}

\end{center}
\caption{Log-log plot of pre-training loss and supervised fine-tuning loss versus the number of non-embedding parameters for the Mini/Small/Medium/Base/Base24 model configurations (see Table \ref{tab:models}) on the PCQM4M-v2 dataset.}
\label{fig:scaling-law-pt-sft}
\end{figure}

The pre-training and fine-tuning configurations for PCQM4M-v2 are in Tab.~\ref{tab:graph-pcqm-hp}. 
The log-scale scaling law plot for both pre-training loss and supervised fine-tuning loss is shown in Fig.~\ref{fig:scaling-law-pt-sft}.

\begin{table}[ht]
\caption{Pre-train and fine-tune configurations for the PCQM4M-v2 dataset. LSI means layer-scale-initialization, EMA is exponential moving average, MPE stands for max-position-embedding, and TWE means tie-word-embeddings.}
\label{tab:graph-pcqm-hp}
\vskip 0.15in
\begin{center}
\begin{small}
\begin{tabular}{p{3cm}|p{5.2cm} p{5.2cm}}
\toprule
                                  & \multicolumn{1}{c}{pre-train}            & \multicolumn{1}{c}{fine-tune}            \\
\midrule
model-size                        & \multicolumn{2}{c}{Mini Small Medium Base Base24 Base48}                     \\
batch-size                        & \multicolumn{1}{c}{$1024/1024/1024/1024/8192/8192$}         & \multicolumn{1}{c}{1024}          \\
total                             & \multicolumn{1}{c}{$1/1/1/1/4/4 \times 10^9$ tokens}        & \multicolumn{1}{c}{32 epochs}         \\
warmup                            & \multicolumn{1}{c}{$10^8$ tokens}                           & \multicolumn{1}{c}{9.6 epochs}         \\

\midrule

lr scheduler                      & \multicolumn{1}{c}{Warmup \& linear decay}    & \multicolumn{1}{c}{Warmup \& cosine decay}     \\
max-lr                            & \multicolumn{1}{c}{$3\times 10^{-4}$}         & \multicolumn{1}{c}{$6/6/6/6/2/1.5 \times 10^{-4}$}           \\
min-lr                            & \multicolumn{1}{c}{0}                         & \multicolumn{1}{c}{automatic set}        \\
Adam-betas                        & \multicolumn{1}{c}{$[0.9, 0.95]$}             & \multicolumn{1}{c}{$[0.9, 0.99]$}       \\
Adam-eps                          & \multicolumn{1}{c}{$1\times 10^{-8}$}         & \multicolumn{1}{c}{$1\times 10^{-10}$}       \\
max-grad-norm                     & \multicolumn{1}{c}{5}                         & \multicolumn{1}{c}{1}        \\
weight-decay                      & \multicolumn{1}{c}{0.1}                       & \multicolumn{1}{c}{0.02}    \\

\midrule

attention-dropout                 & \multicolumn{2}{c}{0.1}                                                      \\
path-dropout                      & \multicolumn{1}{c}{0}                          & \multicolumn{1}{c}{$0/0/0/0.05/0.1/0.2$}    \\
embed-dropout                     & \multicolumn{2}{c}{0}                                                        \\
mlp-dropout                       & \multicolumn{2}{c}{0}                                                        \\
LSI-val                           & \multicolumn{1}{c}{NA}                         & \multicolumn{1}{c}{1}           \\
EMA                               & \multicolumn{2}{c}{NA}                                                      \\

\midrule

hidden-act                                       & \multicolumn{2}{c}{gelu}                                                     \\
MPE                                              & \multicolumn{2}{c}{1024}                                                      \\
TWE                                              & \multicolumn{1}{c}{FALSE}                      & \multicolumn{1}{c}{NA}       \\
\bottomrule
\end{tabular}
\end{small}
\end{center}
\vskip -0.1in
\end{table}

\section{Edge-Level Task}
\label{app:edge}




\subsection{ogbl-ppa}

\begin{figure}[thpb]
\begin{center}
\includegraphics[width=0.6\textwidth]{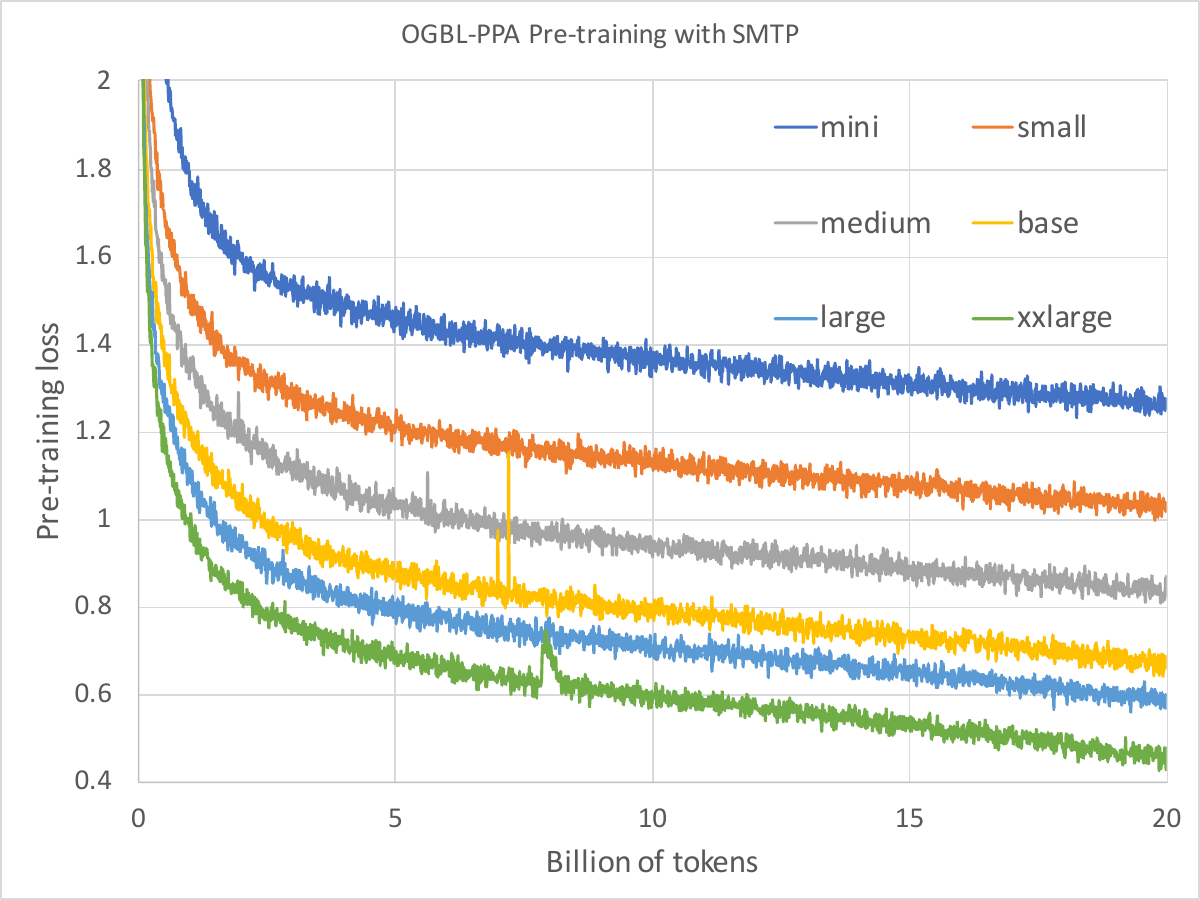}
\end{center}
\caption{Pre-train loss versus tokens of ogbl-ppa dataset for models mini/small/medium/base/large/xxlarge as in Tab. \ref{tab:models}.
}
\label{fig:scaling-ppa}
\end{figure}


We use two tokens for the node identity encoding introduced in Sec. \ref{sec:node-coding}. Specifically, we use the species to partition the nodes, so the first token represents the species, and the second is the local indices of proteins inside each species.

\begin{table}[ht]
\caption{The Pre-training and fine-tuning configurations for the ogbl-ppa dataset. For XXL model, we use fp16 in the pre-train stage, and bf16 in the fine-tune stage for numerical stability.}
\label{tab:edge-ppa-hp}
\vskip 0.15in
\begin{center}
\begin{small}
\begin{tabular}{p{3cm}|p{5.2cm} p{5.2cm}}
\toprule
                        & \multicolumn{1}{c}{pre-train}                     & \multicolumn{1}{c}{fine-tune}                                                \\
\midrule
model-size              & \multicolumn{2}{c}{Mini Small Medium Base Large XXLarge}                     \\
batch-size              & \multicolumn{1}{c}{$1024$}                        & \multicolumn{1}{c}{$8192$}      \\
total                   & \multicolumn{1}{c}{$2 \times 10^{10}$ tokens}     & \multicolumn{1}{c}{$8/8/8/8/16/16$ epochs}     \\
warmup                  & \multicolumn{1}{c}{$10^9$ tokens}                 & \multicolumn{1}{c}{$2.4/2.4/2.4/2.4/4.8/4.8$ epochs} \\

\midrule

lr scheduler            & \multicolumn{1}{c}{Warmup \& linear decay}        & \multicolumn{1}{c}{Warmup \& cosine decay}   \\
max-lr                  & \multicolumn{1}{c}{$3\times 10^{-4}$}             & \multicolumn{1}{c}{$3/3/3/3/3/2 \times 10^{-5}$}    \\
min-lr                  & \multicolumn{1}{c}{$0$}                             & \multicolumn{1}{c}{automatic set}    \\
Adam-betas              & \multicolumn{1}{c}{$[0.9, 0.95]$}                 & \multicolumn{1}{c}{$[0.9, 0.99]$}      \\
Adam-eps                & \multicolumn{1}{c}{$1\times 10^{-8}$}             & \multicolumn{1}{c}{$1\times 10^{-10}$}     \\
max-grad-norm           & \multicolumn{1}{c}{5}                             & \multicolumn{1}{c}{1}                                                        \\
weight-decay            & \multicolumn{1}{c}{0.1}                           & \multicolumn{1}{c}{0}                                                       \\

\midrule

attention-dropout       & \multicolumn{2}{c}{0.1}                                                      \\
path-dropout            & \multicolumn{1}{c}{$0/0/0/0/0.1/0.1$}             & \multicolumn{1}{c}{$0/0/0/0.05/0.1/0.2$}       \\
embed-dropout           & \multicolumn{2}{c}{0}                                                        \\
mlp-dropout             & \multicolumn{2}{c}{0}                                                        \\
LSI-val                 & \multicolumn{1}{c}{NA}                            & \multicolumn{1}{c}{$\text{NA}/\text{NA}/1/1/1/\text{NA}$}          \\
EMA                     & \multicolumn{2}{c}{NA}                                                       \\

\midrule

hidden-act              & \multicolumn{2}{c}{gelu}                                                     \\
MPE                     & \multicolumn{2}{c}{$1024$}                                                     \\
TWE                     & \multicolumn{1}{c}{FALSE}                          & \multicolumn{1}{c}{NA}       \\
\bottomrule
\end{tabular}
\end{small}
\end{center}
\vskip -0.1in
\end{table}

The pre-training and fine-tuning configurations for ogbl-ppa are listed in Tab. \ref{tab:edge-ppa-hp}. The loss of pre-training versus the number of tokens is shown in Fig. \ref{fig:scaling-ppa}. 

The fine-tuning data consists of subgraphs induced by the positive edges for training and equal negative edges randomly sampled. 

In general, a larger model results in lower pre-training loss, and better results in down-stream fine-tuning tasks.

\subsection{ogbl-citation2}

\begin{table}[ht]
\caption{Pre-train and fine-tune configurations for the ogbl-citation2 dataset. We use bf16 in both the pre-training and fine-tuning stages for numerical stability. One epochs contains 10\% randomly sampled positive edges and negative edges. For a given positive edge of head and tail node, we randomly sample a node as the tail node, and then form a negative edge with the head node.}
\label{tab:edge-citation2-hp}
\vskip 0.15in
\begin{center}
\begin{small}
\begin{tabular}{p{3cm}|p{5.2cm} p{5.2cm}}
\toprule
                        & \multicolumn{1}{c}{pre-train}                     & \multicolumn{1}{c}{fine-tune}                                                \\
\midrule
model-size              & \multicolumn{2}{c}{Medium Base}                     \\
batch-size              & \multicolumn{1}{c}{$1024$}                        & \multicolumn{1}{c}{$4096/2048$}      \\
total                   & \multicolumn{1}{c}{$2 \times 10^{10}$ tokens}     & \multicolumn{1}{c}{$32$ epochs}     \\
warmup                  & \multicolumn{1}{c}{$10^9$ tokens}                 & \multicolumn{1}{c}{$9.6$ epochs} \\

\midrule

lr scheduler            & \multicolumn{1}{c}{Warmup \& linear decay}        & \multicolumn{1}{c}{Warmup \& cosine decay}   \\
max-lr                  & \multicolumn{1}{c}{$1\times 10^{-4}$}             & \multicolumn{1}{c}{$3 \times 10^{-5}$}    \\
min-lr                  & \multicolumn{1}{c}{$0$}                           & \multicolumn{1}{c}{automatic set}    \\
Adam-betas              & \multicolumn{1}{c}{$[0.9, 0.95]$}                 & \multicolumn{1}{c}{$[0.9, 0.99]$}      \\
Adam-eps                & \multicolumn{1}{c}{$1\times 10^{-8}$}             & \multicolumn{1}{c}{$1\times 10^{-10}$}     \\
max-grad-norm           & \multicolumn{2}{c}{1}                                                        \\
weight-decay            & \multicolumn{1}{c}{0.1}                           & \multicolumn{1}{c}{0}                                                       \\

\midrule

attention-dropout       & \multicolumn{2}{c}{0.1}                                                      \\
path-dropout            & \multicolumn{1}{c}{$0$}                           & \multicolumn{1}{c}{$0.05$}       \\
embed-dropout           & \multicolumn{2}{c}{0}                                                        \\
mlp-dropout             & \multicolumn{2}{c}{0}                                                        \\
LSI-val                 & \multicolumn{2}{c}{N/A}          \\
EMA                     & \multicolumn{2}{c}{N/A}                                                       \\

\midrule

hidden-act              & \multicolumn{2}{c}{gelu}                                                     \\
MPE                     & \multicolumn{1}{c}{$512$}                          & \multicolumn{1}{c}{$1024$}   \\
TWE                     & \multicolumn{1}{c}{FALSE}                          & \multicolumn{1}{c}{N/A}       \\
\bottomrule
\end{tabular}
\end{small}
\end{center}
\vskip -0.1in
\end{table}

\begin{figure}[thpb]
\begin{center}
\includegraphics[width=0.6\textwidth]{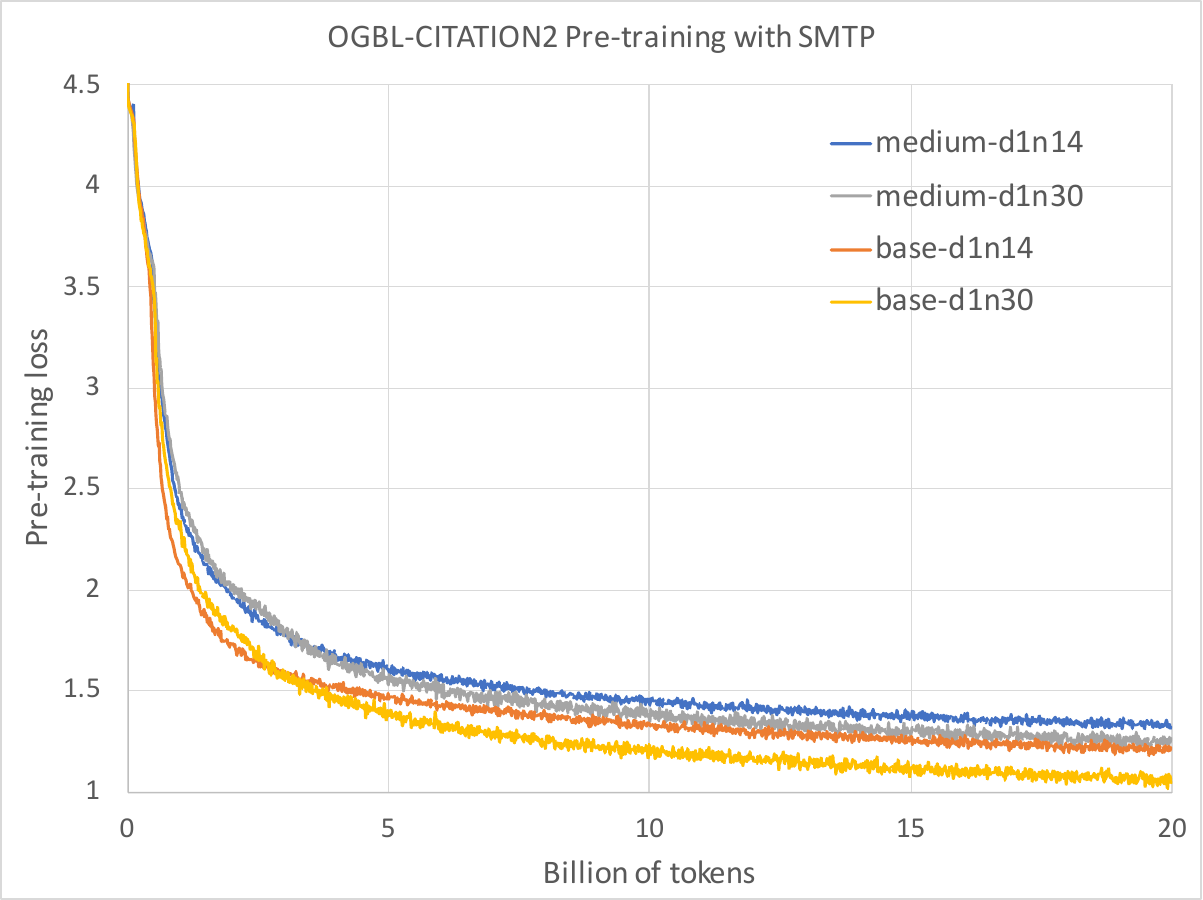}
\end{center}
\caption{Pre-train loss versus tokens of ogbl-citation2 dataset for models medium/base as in Tab. \ref{tab:models}.
}
\label{fig:scaling-citation2}
\end{figure}

The pre-training and fine-tuning configurations for ogbl-citation2 are listed in Tab. \ref{tab:edge-citation2-hp}. The subgraph sampling is edge-ego with $d=1$ and $n=14/30$ as in Tab. \ref{tab:sampling}. The pre-training losses of the ogbl-citation2 dataset with different model sizes and subgraph sampling settings are shown in Fig. \ref{fig:scaling-citation2}.

\section{Node-Level Task}
\label{app:node}

\subsection{ogbn-proteins}

\begin{table}[ht]
\caption{Configurations of pre-training with SMTP and fine-tuning for the ogbn-proteins dataset.}
\label{tab:node-proteins-hp}
\vskip 0.15in
\begin{center}
\begin{small}
\begin{tabular}{p{3cm}|p{5.2cm} p{5.2cm}}
\toprule
                        & \multicolumn{1}{c}{pre-train}                     & \multicolumn{1}{c}{fine-tune}                                                \\
\midrule
model-size              & \multicolumn{2}{c}{Small Medium Base}                     \\
batch-size              & \multicolumn{1}{c}{$256$}                         & \multicolumn{1}{c}{$128$}      \\
total                   & \multicolumn{1}{c}{$2 \times 10^{10}$ tokens}     & \multicolumn{1}{c}{$16/16/8$ epochs}     \\
warmup                  & \multicolumn{1}{c}{$10^9$ tokens}                 & \multicolumn{1}{c}{$4.8/4.8/2.4$ epochs} \\

\midrule

lr scheduler            & \multicolumn{1}{c}{Warmup \& linear decay}        & \multicolumn{1}{c}{Warmup \& cosine decay}   \\
max-lr                  & \multicolumn{1}{c}{$3\times 10^{-4}$}             & \multicolumn{1}{c}{$3 \times 10^{-5}$}    \\
min-lr                  & \multicolumn{1}{c}{$0$}                           & \multicolumn{1}{c}{automatic set}    \\
Adam-betas              & \multicolumn{1}{c}{$[0.9, 0.95]$}                 & \multicolumn{1}{c}{$[0.9, 0.99]$}      \\
Adam-eps                & \multicolumn{1}{c}{$1\times 10^{-8}$}             & \multicolumn{1}{c}{$1\times 10^{-10}$}     \\
max-grad-norm           & \multicolumn{2}{c}{1}                                                        \\
weight-decay            & \multicolumn{1}{c}{0.1}                           & \multicolumn{1}{c}{0}                                                       \\

\midrule

attention-dropout       & \multicolumn{2}{c}{0.1}                                                      \\
path-dropout            & \multicolumn{2}{c}{0}                                  \\
embed-dropout           & \multicolumn{1}{c}{0}                             & \multicolumn{1}{c}{$0.1/0.2/0.1$}                           \\
mlp-dropout             & \multicolumn{2}{c}{0}                                                        \\
LSI-val                 & \multicolumn{2}{c}{N/A}          \\
EMA                     & \multicolumn{1}{c}{N/A}                           & \multicolumn{1}{c}{$0.999$} \\

\midrule

hidden-act              & \multicolumn{2}{c}{gelu}                                                     \\
MPE                     & \multicolumn{2}{c}{$512$}   \\
TWE                     & \multicolumn{1}{c}{FALSE}                          & \multicolumn{1}{c}{N/A}       \\
\bottomrule
\end{tabular}
\end{small}
\end{center}
\vskip -0.1in
\end{table}

The configurations of pre-training and fine-tuning are in Tab. \ref{tab:node-proteins-hp}. The subgraph sampling is node-ego with $d=20$ and $n=1$ as in Tab. \ref{tab:sampling}. The node identity is encoded with two tokens similar to the ogbl-ppa in Sec. \ref{sec:exp-edgelevel} (see App. \ref{app:edge} for details).

\subsection{ogbn-arxiv}

\begin{table}[ht]
\caption{Configurations of pre-training with SMTP and fine-tuning for the ogbn-arxiv dataset.}
\label{tab:node-arxiv-hp}
\vskip 0.15in
\begin{center}
\begin{small}
\begin{tabular}{p{3cm}|p{5.2cm} p{5.2cm}}
\toprule
                        & \multicolumn{1}{c}{pre-train}                     & \multicolumn{1}{c}{fine-tune}                                                \\
\midrule
model-size              & \multicolumn{2}{c}{Small Medium Base}                     \\
batch-size              & \multicolumn{1}{c}{$256$}                         & \multicolumn{1}{c}{$128$}      \\
total                   & \multicolumn{1}{c}{$4 \times 10^{9}$ tokens}     & \multicolumn{1}{c}{$4$ epochs}     \\
warmup                  & \multicolumn{1}{c}{$2 \times 10^8$ tokens}                 & \multicolumn{1}{c}{$1.2$ epochs} \\

\midrule

lr scheduler            & \multicolumn{1}{c}{Warmup \& linear decay}        & \multicolumn{1}{c}{Warmup \& cosine decay}   \\
max-lr                  & \multicolumn{1}{c}{$3\times 10^{-4}$}             & \multicolumn{1}{c}{$3/3/2 \times 10^{-4}$}    \\
min-lr                  & \multicolumn{1}{c}{$0$}                           & \multicolumn{1}{c}{automatic set}    \\
Adam-betas              & \multicolumn{1}{c}{$[0.9, 0.95]$}                 & \multicolumn{1}{c}{$[0.9, 0.99]$}      \\
Adam-eps                & \multicolumn{1}{c}{$1\times 10^{-8}$}             & \multicolumn{1}{c}{$1\times 10^{-10}$}     \\
max-grad-norm           & \multicolumn{2}{c}{1}                                                        \\
weight-decay            & \multicolumn{1}{c}{0.1}                           & \multicolumn{1}{c}{0}                                                       \\

\midrule

attention-dropout       & \multicolumn{2}{c}{0.1}                                                      \\
path-dropout            & \multicolumn{1}{c}{0}                             & \multicolumn{1}{c}{$0/0/0.1$}     \\
embed-dropout           & \multicolumn{1}{c}{0}                             & \multicolumn{1}{c}{$0.1$}                           \\
mlp-dropout             & \multicolumn{2}{c}{0}                                                        \\
LSI-val                 & \multicolumn{2}{c}{N/A}          \\
EMA                     & \multicolumn{1}{c}{N/A}                           & \multicolumn{1}{c}{$0.9997$} \\

\midrule

hidden-act              & \multicolumn{2}{c}{gelu}                                                     \\
MPE                     & \multicolumn{2}{c}{$1024$}   \\
TWE                     & \multicolumn{1}{c}{FALSE}                          & \multicolumn{1}{c}{N/A}       \\
\bottomrule
\end{tabular}
\end{small}
\end{center}
\vskip -0.1in
\end{table}

The configurations of pre-training and fine-tuning are in Tab. \ref{tab:node-arxiv-hp}. The subgraph sampling is node-ego with $d=1$ and $n=40$ as in Tab. \ref{tab:sampling}. The node identity is encoded with two tokens.

\section{Question and Answering}

\textbf{Q1. Evaluating GraphGPT on real-world citation networks (e.g., PubMed, Cora) or social networks (e.g., Twitter, Facebook graphs) could be great.}

\textit{A1.} We evaluated GraphGPT on large-scale real-world citation networks: ogbn-arxiv (169K nodes, 1.17M edges) and ogbl-citation2 (2.93M nodes, 30.6M edges). These datasets are significantly larger than traditional benchmarks like Cora (2.7K nodes, 5.4K edges) and PubMed (19.7K nodes, 44.3K edges), aligning with our focus on scaling to massive graph data.

We chose these datasets because GraphGPT’s performance benefits from large-scale pre-training data to learn inductive biases (e.g., node permutation invariance). For instance, pre-training on the small Triangles dataset (45K graphs) yielded poor fine-tuning results (32.6\%), whereas scaling pre-training data improved performance to 99\% (Section \ref{sec:exp-graph-struct}). This mirrors the trend in Vision Transformers (ViT), which outperform CNNs only with sufficiently large datasets \citep{dosovitskiy2021image}.

While GNNs may outperform GraphGPT on small datasets like Cora or PubMed, our goal is to demonstrate scalability for large-scale graphs—a critical challenge in modern applications.

\textbf{Q2. While GraphGPT enables a lossless and reversible graph-to-seq transformation, how well does it do this in real-world noisy graphs?}

\textit{A2.} While not the focus of this paper, we tested GraphGPT on an internal noisy graph dataset (3.1M graphs, avg. 24.8 nodes, 54.7 edges) for edge denoising.

Using a semi-supervised node classification task, GraphGPT achieved 10-20\% F1 score improvement over baselines. We formulated the task analogously to Part-of-Speech tagging in NLP, leveraging token-level embeddings. The `long' variant outperformed `short' (see Fig. \ref{fig:eulerize}) likely due to its edge-agnostic token embeddings of nodes.

Results were robust enough for online deployment.

\textbf{Q3. What are the run-time comparisons with GNNs?}

\textit{A3.} We evaluated the run-time of GraphGPT versus GNNs using the PCQM4M-v2 dataset on a single V100-32G GPU. The GNN baselines (adopted from \citet{hu2021ogb}) were implemented using the official GitHub repository\footnote{\url{https://github.com/snap-stanford/ogb/tree/master/examples/lsc/pcqm4m-v2}}.

Results are shown in Table \ref{tab:runtime}. Run-time remains nearly constant across GraphGPT models ranging from 0.62M to 33.95M parameters. This consistency stems from an IO bottleneck during CPU-based data preprocessing. This time-consuming preprocessing phase involves determining if a graph is Eulerian, Eulerizing non-Eulerian graphs, and generating Eulerian paths.

Overall, GraphGPT's run-time is comparable to GNNs when model sizes are similar.

\begin{table}[ht]
\caption{Run-time comparison between GraphGPT variants and GNNs on the PCQM4Mv2 dataset. Time per epoch is measured in minutes.}
\label{tab:runtime}
\vskip 0.15in
\begin{center}
\begin{small}
\begin{tabular}{l|l|l}
model           & \# params & time per epoch \\
\toprule
GIN             & 3.76 M    & 9.25 min       \\
GIN-virtual     & 6.66 M    & 11.2 min       \\
GCN             & 1.96 M    & 8.0 min        \\
GCN-virtual     & 4.85 M    & 9.6 min        \\
\midrule
GraphGPT-Tiny   & 0.62 M    & 20 min         \\
GraphGPT-Mini   & 4.39 M    & 21 min         \\
GraphGPT-Small  & 17.17 M   & 20 min         \\
GraphGPT-Medium & 33.95 M   & 20 min        \\
GraphGPT-Base   & 113.85 M  & 46.7 min        \\
\bottomrule
\end{tabular}
\end{small}
\end{center}
\vskip -0.1in
\end{table}

\textbf{Q4. What's the computational cost of GraphGPT models?}

\textit{A4.} We have included computational cost details for the PCQM4Mv2 dataset in \S \ref{app:discussion}. The cost of other datasets are in the Tab. \ref{tab:cost}.

\begin{table}[ht]
\caption{Computational cost details of the main datasets in the paper. `PT' means pre-training and `FT' stands for fine-tuning. Time is measured in hours. The model size is `Base' as in Tab. \ref{tab:models} with number of parameters about 110M. The corresponding hyper-parameters can be found in Tab. \ref{tab:edge-ppa-hp}, \ref{tab:edge-citation2-hp}, \ref{tab:node-proteins-hp}, \ref{tab:node-arxiv-hp}.}
\label{tab:cost}
\vskip 0.15in
\begin{center}
\begin{small}
\begin{tabular}{l|l|l|l|l|l}
\toprule
dataset        & model size & PT time & FT time  & GPU-PT       & GPU-FT       \\
\midrule
ogbl-ppa       & B          & 58.73 h & 112.62 h & 8 Nvidia L20 & 16 V100-32G  \\
ogbl-citation2 & B          & 72 h    & 100.3 h  & 8 Nvidia L20 & 8 Nvidia L20 \\
\midrule
ogbn-proteins  & B          & 27.1 h  & 3.1 h    & 8 Nvidia L20 & 1 V100-32G   \\
ogbn-arxiv     & B          & 9.25 h  & 4.3 h    & 8 Nvidia L20 & 1 V100-32G   \\
\bottomrule
\end{tabular}
\end{small}
\end{center}
\vskip -0.1in
\end{table}


\textbf{Q5. How robust is the model to adversarial graph perturbations?}

\textit{A5.} Adversarial robustness is a promising research area across NLP, CV, and graphs \citep{guo2021adversarial,shao2022adversarial,jin2020adversarial,sun2023adversarial}. While not our primary focus, preliminary results on noisy graphs (\textbf{Q2}) suggest robustness through large-scale training. A deeper study would bridge GraphGPT’s transformer architecture with adversarial graph defenses, an encouraging future direction.

\textbf{Q6. Can GraphGPT generate graphs that match real-world constraints (e.g., chemical validity)?}

\textit{A6.} While generation is not the primary focus, preliminary experiments show GraphGPT can generate valid molecules after pre-trained on PCQM4M-v2.

However, generation quality depends on hyperparameters (e.g., temperature, top-p, iteration count T). Unconditional/conditional generation and diversity control require further study, which is planned for future work.

\textbf{Q7. Can you compare with other pre-trained-based graph models to highlight the advantages of GraphGPT?}

\textit{A7.} While models like GraphBERT \citep{zhang2020graphbert}, GraphMAE \citep{hou2022graphmae}, and GCC \citep{qiu2020gcc} employ graph pre-training, they primarily target small-scale datasets. GraphGPT’s evaluation focuses on large-scale OGB leaderboard benchmarks, where existing pre-trained models lack competitive entries. Our comparisons align with state-of-the-art baselines dominating these leaderboards, emphasizing scalability and performance on real-world graph tasks.

\textbf{Q8. Why there is a need to introduce stochasticity when doing path identification?}

\textit{A8.} GraphGPT lacks some inductive biases inherent to GNNs (e.g., node permutation invariance). Randomly sampling Eulerian paths per epoch forces the model to learn invariance between different paths of the same graph, akin to how ViT (lacking CNN’s inductive biases) benefits from large-scale data and data augmentation. Empirically, this reduced overfitting on molecular datasets.

\textbf{Q9. Size to performance ratio.}

\textit{A9.} We clarify parameter counts and performance across datasets below:

Graph-Level (Tab. 1–2): GraphGPT’s parameter counts are comparable to prior SOTA (e.g., 113.6M vs. 86M for GPTrans).

Edge-Level (Tab.4): For ogbl-ppa, GraphGPT-B (145.3M) is a bit worse than Refined-GAE (295.8M), but GraphGPT-XXL (2B) achieves the highest performance. For ogbl-citation2, GraphGPT-M (46.8M) and GraphGPT-B (133.1M) outperform MPLP (749.8M).

Node-Level (Tab.5): GraphGPT requires larger parameters on ogbn-proteins and ogbn-arxiv. This may reflect insufficient pre-training data for these tasks, leading to suboptimal parameter utilization.

\textbf{Q10. It is unclear what is intended by model scalability in \S \ref{sec:exp-edgelevel}; additionally scalability seems to not be the answer to the problem if we take into account costs and computational resources required to solve the tasks.}

\textit{A10.} Our investigation of model scalability serves two critical purposes:

1. Studying performance limits reveals fundamental insights of data. Even small performance gains can reduce real-world validation costs \citep{ogbl-ppa-web}.

2. This study aligns with foundational NLP scaling law research \citep{kaplan2020scaling, hoffmann2022training}, aiming to catalyze similar investigations for graph-structured data.

\textbf{Q11. How do you evaluate the correctness of the response? Do you query the Transformer model again with additional information if the response is not correct?}

\textit{A11.} The model directly outputs predictions via the task head during inference. Results are evaluated using standard metrics (e.g., MAE, accuracy) for the downstream task. Each test/valid instance is processed once; no iterative querying is performed.

\textbf{Q12. How is the prompt structured? How do you express the task to solve?}

\textit{A12.} We do not use prompts. Instead, tasks are encoded via specialized tokens appended to the input sequence and processed by an additional MLP head during fine-tuning as discussed in \S \ref{sec:graph-tasks}. Fig. \ref{fig:pre-train-fine-tune} illustrates the implementations.


\end{document}